\documentclass[letterpaper, 10 pt, journal, twoside]{ieeetran} 
\IEEEoverridecommandlockouts                                                                

\usepackage{comment}
\usepackage{graphicx} 
\usepackage{amsmath} 
\usepackage{amssymb}  
\usepackage{cite}
\usepackage{xcolor}
\usepackage[hidelinks]{hyperref}
\usepackage[caption=false, font=scriptsize, labelfont=sf, textfont=sf]{subfig}
\usepackage{bm}
\usepackage{adjustbox}
\usepackage{placeins}
\usepackage{soul}
\usepackage[font={small}]{caption}

\title{\LARGE \bf
A Combined Learning and Optimization Framework to Transfer Human Whole-body Loco-manipulation Skills to Mobile Manipulators

}

\author{Jianzhuang Zhao$^{1,2}$, Francesco Tassi$^{1}$, Yanlong Huang$^{3}$,
Elena De Momi$^{2}$, and Arash Ajoudani$^{1}$
\thanks{1- Human-Robot Interfaces and Interaction Lab, Istituto Italiano di Tecnologia, Genoa, Italy. 2- Dept. of Electronics, Information, and Bioengineering, Politecnico di Milano, Italy. 3- School of Computing, University of Leeds, Leeds LS29JT, UK.}
\thanks{This work was supported by the European Research Council's (ERC) starting grant Ergo-Lean (GA 850932).}
\thanks{E-mail: \tt\small jianzhuang.zhao@iit.it}
}

\begin{document}

\maketitle
\thispagestyle{empty}
\pagestyle{empty}

\begin{abstract}
Humans' ability to smoothly switch between locomotion and manipulation is a remarkable feature of sensorimotor coordination. Leaning and replication of such human-like strategies can lead to the development of more sophisticated robots capable of performing complex whole-body tasks in real-world environments. To this end, this paper proposes a combined learning and optimization framework for transferring human's loco-manipulation soft-switching skills to mobile manipulators. The methodology departs from data collection of human demonstrations for a locomotion-integrated manipulation task through a vision system. Next, the wrist and pelvis motions are mapped to mobile manipulators' End-Effector (EE) and mobile base. A kernelized movement primitive algorithm learns the wrist and pelvis trajectories and generalizes to new desired points according to task requirements. Next, the reference trajectories are sent to a hierarchical quadratic programming controller, where the EE and the mobile base reference trajectories are provided as the first and second priority tasks, generating the feasible and optimal joint level commands. A locomotion-integrated pick-and-place task is executed to validate the proposed approach. After a human demonstrates the task, a mobile manipulator executes the task with the same and new settings, grasping a bottle at non-zero velocity. The results showed that the proposed approach successfully transfers the human loco-manipulation skills to mobile manipulators, even with different geometry.  

\end{abstract}


\section{Introduction}
In recent years, the integration of Mobile Manipulators (MM) has emerged as a pivotal paradigm in robotics, representing a transformative leap in the capabilities of autonomous systems. Combining the mobility of mobile platforms with the dexterity of manipulator arms, MMs possess a unique versatility that enables unprecedented possibilities in various domains, ranging from manufacturing and logistics to healthcare and disaster response~\cite{roa2021mobile, Zhao2022}. However, due to the high redundancy caused by the combination of these two components, the complexity of MMs' trajectory planning and motion control noticeably increases~\cite{sandakalum2022motion}. 

The planning methods for MMs can be divided into two classes, depending on the treatment of the two components, i.e., two separated subsystems or one whole-body system~\cite{sandakalum2022motion}. In the first case, the mobile platform moves to a location close to the target object (locomotion mode). Then, the robotic arm moves to manipulate the object (manipulation mode). In this manner, the trajectory can be planned separately. For example, in~\cite{castaman2021receding}, the trajectory of the mobile platform is planned by the Djikstra algorithm, and the trajectory of the robotic arm is generated by RRTConnect later. One important issue of this approach is where to locate the mobile platform to ensure the reachability of the robotic arm in the following manipulation mode since the robotic arm cannot reach the object if the base is too far. To address this issue, the Inverse Reachability Map (IRM) is proposed in~\cite{vahrenkamp2013robot}. Specifically, for a given end-effector's (EE) pose of the MM, the proper poses of the mobile base and feasible configurations of the robotic arm can be found by querying IRM. Recently, the IRM has been applied as priors in the reinforcement learning (RL) method in~\cite{jauhri2022robot} to accelerate the learning process. Although IRM can be generated offline and searched online, its construction is non-trivial. Most importantly, IRM is specific to each robot and cannot be generalized to another one. The planning methods of the two sub-systems result in separate and sequential movements of the mobile base and robotic arm, which decrease task efficiency and inhibit the optimal execution of tasks requiring complex and coordinated motions of the two components (e.g., doors opening).

\begin{figure}[t]
    \centering 
    \includegraphics[trim=6.4cm 2.9cm 5.5cm 3.8cm,clip,width=\columnwidth]{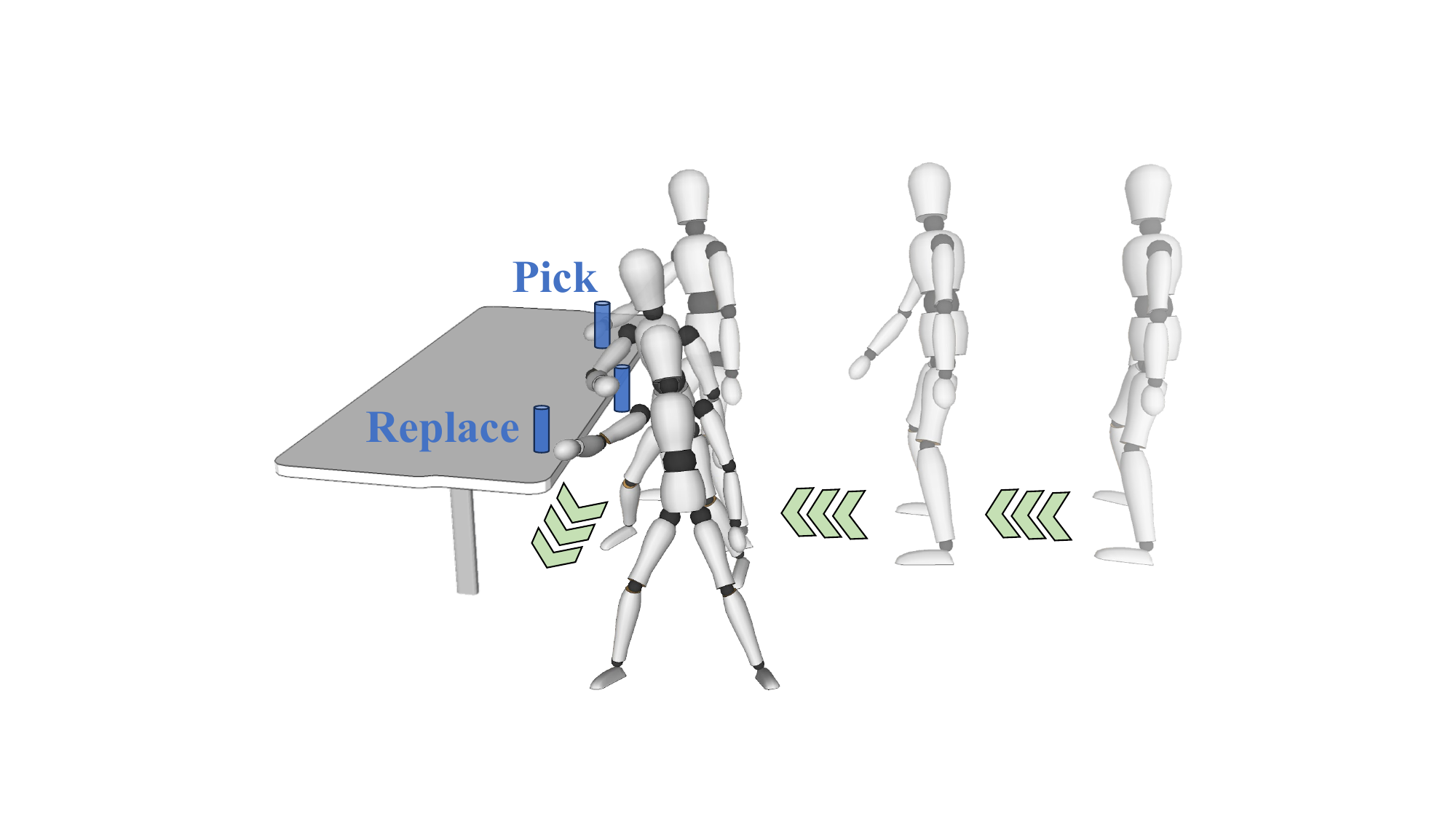} 
    \caption{Human whole-body motion behavior for a locomotion-integrated pick-and-place task: the blue target object is outside the workspace of the right arm. Thus, the human first approaches the object and subsequently manipulates it in a single smooth motion.  
    }
    \label{fig:con_fig}
\end{figure}

To improve efficiency and achieve complex tasks, some planning approaches treat MMs as a unique redundant system, generating whole-body trajectories. 
In this way, whole-body trajectory planning can be formulated as an optimization problem, considering the joint limits of the robotic arm and mobile base velocity limits to minimize the predefined cost functions. For instance, an optimization approach is designed in~\cite{zimmermann2021go} to generate whole-body trajectories for the MM that integrates a Kinova robotic arm on a Spot.  
However, the computational burden is high, especially for a high Degree of Freedom (DoF) MM, which often cannot meet real-time requirements. In~\cite{honerkamp2021learning}, the mobile manipulation tasks are divided into an EE planner and a conditional RL agent for the mobile platform. Specifically, the velocity of the mobile platform can be inferred from a given EE velocity to ensure the kinematic feasibility of the robotic arm. Although this method can generate coordinated motion (loco-manipulation mode) for the mobile base and robotic arm, it still needs a standard inverse kinematic solver for joint-level commands. Besides, a whole-body loco-manipulation planning algorithm for a humanoid NAO is proposed in~\cite{ferrari2017humanoid}, where the three motion modes, locomotion, loco-manipulation, and manipulation, are switched at predefined points based on the distance (from far to close) between the robot's Center of Mass (CoM) and the target object. Although it generates natural motion for NAO, the overall motion combines three separate sub-trajectories, and finding the proper switching points is complex.  
More recently, a reactive mobile manipulation architecture has been proposed in~\cite{icra2023onthemove}, which can grasp static or dynamic objects while the mobile base is still moving.

However, most of the aforementioned planning approaches do not consider the impact when the EE comes into contact with the objects, which is essential when the contact velocity is non-zero, namely dynamic manipulation tasks~\cite{learning2022fixbase, zhao2023impact}. For example, for the pick-and-place tasks (static target objects), most approaches use almost zero contact velocity to reduce the impact, generating a stable grasp, which decreases the task efficiency. On the other hand, the target objects may topple over or bounce if the velocity of EE is high at the moment of impact. Although the methods in~\cite{zimmermann2021go, icra2023onthemove} can succeed in grasping without stopping, the objects are grasped from the top, thus not contributing to tilting the object. 
Therefore, choosing a proper contact velocity when dynamically grasping objects from the radial direction is still challenging~\cite{learning2022fixbase}. Humans are experts in dynamic manipulation tasks, and we tend to pick objects in a single, fluent, fast movement. Inspired by this, in~\cite{learning2022fixbase}, the fixed-base robotic arm learns to pick static objects at non-zero velocity from humans' interactive demonstrations. Besides, the EE trajectory and contact force are learned for a table cleaning task by kinesthetic teaching in~\cite{Zhao2022}, where the overall work region is small, and the mobile base does not need to move much. In~\cite{lfdmm2017tim}, the proposed approach can learn both the mobile base and end-effector motion from human demonstrations. However, to ensure motion feasibility, the IRM is still used as a constraint to generate the base movement, and the impact is ignored. Therefore, how to learn the whole-body dynamic motion for MMs when the tasks' work region is large is still an open question. 

In some planning/learning approaches, proper controllers should be chosen to distribute the EE motion to the mobile base and robotic arm since only the EE trajectory is defined. Whole-body controllers have been presented to avoid the synchronization issues of decoupled control strategies. A weighted whole-body cartesian impedance controller is proposed in~\cite{wu2021unified} to change the predefined motion modes (i.e., locomotion, manipulation, and loco-manipulation) at given points by setting different weights. This controller can achieve motion mode transfer freely, but a static transition phase is needed to ensure smooth changes between different modes. Furthermore, the robotic arm joint limits and mobile base velocity/acceleration limits are not considered. On the other hand, the Hierarchical Quadratic Programming (HQP) controller in~\cite{tassi2022sociable}
sets the limits as explicit constraints. For a given EE pose at each time step, the HQP controller can ensure the generation of a feasible whole-body command at the joint level.

To address these issues, we propose a combined learning and optimization framework to transfer whole-body human loco-manipulation skills to MMs (see Fig. \ref{fig:con_fig}). Our goal is to generate fluent behaviors for MMs, which can smoothly handle the target objects without stopping the mobile base or robotic arm, thus enabling non-zero contact velocity grasping.
The main contributions of this paper are as follows:
\begin{itemize}
    \item An overall combined learning and optimization framework to achieve locomotion-integrated manipulation tasks for mobile manipulators with any geometry;
    \item A whole-body reference trajectory learning from human demonstrations and generalization formulation;
    \item An HQP controller to transfer the learned whole-body trajectory to joint-level commands and ensure feasibility at each time step. 
\end{itemize}

The following of the paper is organized as follows. The overall framework is described in Sec. \ref{sec:framwork} followed by the details of the proposed methodology in Sec. \ref{sec:methodology}. The experiments and results are presented in \ref{sec:exp}, and Sec. \ref{sec:conclusion} concludes the overall paper. 

\begin{figure*}[t]
	\centering
	\includegraphics[trim=0.75cm 1.15cm 0.75cm 0.8cm,clip,width=.98\linewidth]{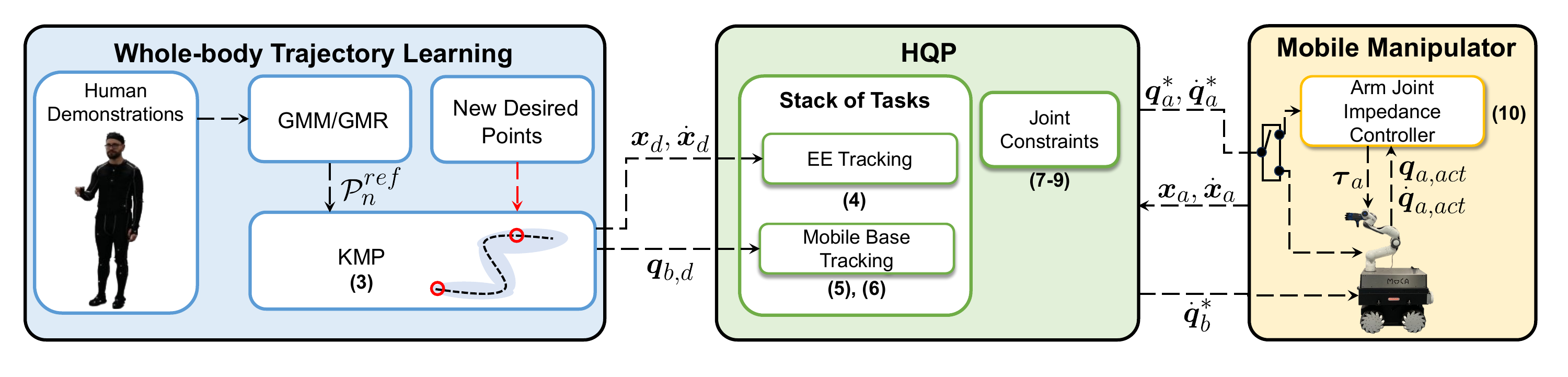}
	\caption{Overall proposed framework. From left to right, inside the whole-body trajectory learning module, the reference trajectories are extracted from human demonstration by GMMs/GMR. After the new desired points are given, KMP generalizes the learned skills to new settings. Next, the desired EE pose, velocity, and mobile base pose are sent to the HQP controller, which computes the optimal joint trajectories based on the hierarchical SoT and imposed constraints. These are passed to the robotic arm's lower-level joint impedance controller and to the mobile base velocity controller. The corresponding equations are also presented in this figure.}
	\label{fig:framework}
\end{figure*}

\section{Overall Framework}\label{sec:framwork}
The overall proposed framework can be found in Fig. \ref{fig:framework}, and comprises two main modules: the whole-body trajectory learning and generalization and the HQP controller. Instead of learning each joint's trajectory for humans, we treat the wrist pose/velocity and pelvis pose as the whole-body trajectory and map them onto both the EE and the mobile base frames, respectively. The Kernelized Movement Primitive (KMP) is applied because of its capacity to generalize to new desired points according to the task requirements. Next, the learned reference trajectories are sent to the HQP controller, where EE and mobile base trajectory tracking tasks are defined as first and secondary priorities to generate the whole-body feasible and optimal joint velocities at each time step. These are sent to the joint impedance controller of the robotic arm to obtain the desired torques. Meanwhile, the mobile base optimal velocity command is sent directly to the default velocity controller.  

\section{Methodology}\label{sec:methodology}

\subsection{Whole-body Trajectory Learning and Generalization}
\label{sec:tra_learn}
\subsubsection{Whole-body Trajectory Mapping}
\label{subsec:maping}
To learn the whole-body motion from human demonstrations, we map the human wrist and pelvis to the mobile manipulators' EE and mobile base, respectively. Since the geometry of the human arm and the mounting method of the robotic arm are different, there is no need to map each human arm joint to the robotic arm. Therefore, the human wrist pose and velocity ($\bm{x}_{wr}$, $\dot{\bm{x}}_{wr}$) $\in\mathbb{R}^{6}$, and pelvis pose $\bm{x}_{pe}$ $\in\mathbb{R}^{6}$, in the world frame are treated as the whole-body trajectories for the MM.

\subsubsection{Trajecoty Learning and Generalization}
\label{subsec:kmp}
in this paper, we implement KMP~\cite{huang2019kernelized} to learn the whole-body trajectory from human demonstrations and adapt to new targets (i.e., initial/via/final points pose and velocity) according to the tasks' requirements.
For the collected $M$ human demonstrations $\{\{ \bm{s}_{n,m}, \bm{\xi}_{n,m}\}_{n=1}^{N_{m}}\}_{m=1}^M$, where we choose each time step as input variable $\bm{s}$ and human wrist pose and velocity, and pelvis pose as output variables $\bm{\xi}=[\bm{x}_{wr}^T \quad \dot{\bm{x}}_{wr}^T \quad \bm{x}_{pe}^T]^T$ for each data point.
Based on these demonstrations, a distribution $\hat{\bm{\xi}}_n|\bm{s}_n\sim\mathcal{N}(\boldsymbol{\hat{\mu}}_n,\boldsymbol{\hat{\Sigma}}_n)$ and the reference trajectory $\{\bm{s}_n,\boldsymbol{\hat{\mu}}_n,\boldsymbol{\hat{\Sigma}}_n\}_{n=1}^N$ can be extracted by  Gaussian Mixture Model (GMMs)/Gaussian Mixture Regression (GMR), where $\boldsymbol{\hat{\mu}}_n$ and $\boldsymbol{\hat{\Sigma}}_n$ represent the mean and covariance, respectively. Then, the parametric model of KMP is written as:
\begin{equation}
	\boldsymbol{\xi}(\boldsymbol{s}) =\boldsymbol{\Phi}^{T}(\boldsymbol{s})\boldsymbol{w},
	\label{eq:kmp}
\end{equation}
where $\boldsymbol{\Phi}(\boldsymbol{s})=\boldsymbol{I}_\mathcal{O}\otimes\boldsymbol{\phi}(\boldsymbol{s})\in\mathbb{R}^{B\mathcal{O}\times\mathcal{O}}$, $\boldsymbol{\phi}(\boldsymbol{s})$ is $B$-dimensional basis function and $\boldsymbol{I}_\mathcal{O}$ is the $\mathcal{O}$ dimensional identity matrix. The weight vector is $\boldsymbol{w}\sim\mathcal{N}(\boldsymbol{{\mu}}_w,\boldsymbol{{\Sigma}}_w)$, where $\boldsymbol{{\mu}}_w$ and $\boldsymbol{{\Sigma}}_w$ are unknown. To obtain these variables, KMP uses \eqref{eq:kl} to minimize the KL-divergence between the probabilistic trajectory generated by \eqref{eq:kmp} and the reference trajectory $\mathcal{P}_n^{ref}$ that models the distribution of demonstrations:
\begin{equation}\label{eq:kl}
	\sum_{n=1}^{N}{KL}(\mathcal{P}_n^{para}||\mathcal{P}_n^{ref}),
\end{equation}
where $\mathcal{P}_n^{para}=\mathcal{N}(\boldsymbol{\Phi}^{T}(\boldsymbol{s}_n)\boldsymbol{{\mu}}_w,\boldsymbol{\Phi}^{T}(\boldsymbol{s}_n)\boldsymbol{\Sigma}_w\boldsymbol{\Phi}(\boldsymbol{s}_n))$ and $\mathcal{P}_n^{ref} = \mathcal{N}(\boldsymbol{\hat{\mu}}_n,\boldsymbol{\hat{\Sigma}}_n)$. By decomposing the above objective function, for any input $s^*$, the corresponding output mean is:

\begin{equation}\label{eq:klmean}
\mathbb{E}(\boldsymbol{\xi}(\boldsymbol{s}^*))=\boldsymbol{k}^*(\boldsymbol{K}+{\lambda}\boldsymbol{\Sigma})^{-1}\boldsymbol{\mu}
\end{equation}
where $\lambda>0$ is regularization factor, $\boldsymbol{k}^*\in\mathbb{R}^{\mathcal{O}\times N\mathcal{O}}$ is a $1 \times N$ block matrix, where the $i$-th column element is $\boldsymbol{k}(\boldsymbol{s}^*,\boldsymbol{s}_i)\boldsymbol{I}_\mathcal{O}$. $\boldsymbol{K}\in\mathbb{R}^{N\mathcal{O}\times N\mathcal{O}}$ is a $N\times N$ block matrix, where the $i$-th row and $j$-th column item is $\boldsymbol{k}(\boldsymbol{s}_i,\boldsymbol{s}_j)\boldsymbol{I}_\mathcal{O}$. Besides, $\boldsymbol{{\mu}}=[\boldsymbol{\hat{\mu}}_1^T\quad \boldsymbol{\hat{\mu}}_2^T\quad...\quad\boldsymbol{\hat{\mu}}_N^T]^T$ and $\boldsymbol{{\Sigma}}=blockdiag \{ {\hat{\boldsymbol{\Sigma}}_1},{\hat{\boldsymbol{\Sigma}}_2},...,{\hat{\boldsymbol{\Sigma}}_N}\}$.
To reproduce learned skills, desired trajectories can be generated directly by \eqref{eq:klmean}. Besides, according to the task requirements, for example, the new robot's initial pose and the new pose of the target object in the pick-and-place task, the learned trajectories must be adapted to the new initial/via pose (i.e., new desired points set  $\{\overline{\bm{s}}_m,\overline{\bm{\mu}}_m,\overline{\bm{\Sigma}}_m\}_{m=1}^M$ ). In this case, the new desired point set can be inserted into the reference trajectory, resulting in a $N+M$ new trajectory. Eventually, \eqref{eq:klmean} is applied to learn and generate the new trajectory, including all the desired points.

\subsection{Hierarchical Quadratic Programming}
\label{subsec:HQP_formulation}
\subsubsection{Formulation} We formulate a Stack of Tasks (SoT) to be solved using an HQP-based control problem as in~\cite{tassi2021augmented}, 
through the establishment of a strict hierarchy. Herein, $N_p\in\mathbb{N}$ denotes the total number of hierarchical layers within the SoT, and $k \in \{1, \dots, N_p\}$ represents the priority levels. These priority levels are arranged in descending order of importance from $k=1$ (most important) to $k=N_p$ (least important). The design ensures that the solution obtained at level $k$ is rigorously imposed on the subsequent lower priority $k+1$. This strict hierarchy leverages the optimality conditions defined at each Quadratic Programming (QP) solution, as elucidated in~\cite{kanoun2011kinematic}.

Specifically, the whole-body Closed-loop Inverse Kinematics (CLIK) of the MM is defined as the first task in the SoT to track both the learned Cartesian EE's pose and velocity $\bm{x}_d, \dot{\bm{x}}_d \in \mathbb{R}^6$ (i.e., the $\bm{x}_{wr}$, $\dot{\bm{x}}_{wr}$ in Sec.\ref{sec:tra_learn}). Since the limits of the robotic arm and mobile base are set as constraints of the HQP, the obtained optimal solution for the MM will be feasible at each time step. Meanwhile, the learned mobile base's pose $\bm{q}_{b,d} \in \mathbb{R}^{n_b}$  (i.e., the $\bm{x}_{pe}$ in Sec.\ref{sec:tra_learn}) tracking is treated as the secondary objective to solve the redundancy of the MM. In this way, the learned whole-body loco-manipulation trajectories are transferred from human to robot, where the tasks are achieved by tracking the EE's pose and velocity in Cartesian space (for example, pick and place at non-zero velocity) strictly and following the learned mobile pose as much as possible, simultaneously. The details of the HQP formulation are presented in what follows.

\subsubsection{Closed-loop Inverse Kinematics}
as mentioned above, the first task in the SoT is the CLIK, which closes the loop on the Cartesian EE error to achieve trajectory tracking between the learned desired pose and the velocity trajectories $\bm{x}_d, \dot{\bm{x}}_d$:
\begin{equation} \label{eq:min_kyn_clik}
    \min_{\dot{\bm{q}}} \,\| \bm{J}\dot{\bm{q}}-( \bm{K_v} ( \dot{\bm{x}}_d -\dot{\bm{x}}_a ) + \bm{K_p}\,(\bm{x}_d-\bm{x}_a) ) \|^2,
\end{equation}
where $\bm{x}_a, \,\bm{x}_d \in \mathbb{R}^6$ are the actual and desired EE trajectories, respectively; $\dot{\bm{x}}_a, \,\dot{\bm{x}}_d \in \mathbb{R}^6$ are the actual and desired velocities respectively, while $\bm{K_p}, \bm{K_v} \in \mathbb{R}^{6\times 6}$ are the positive-definite diagonal gain matrices responsible for the accuracy of position and velocity trajectories' tracking.
$\dot{\bm{q}} \in \mathbb{R}^n$ are the joint velocities, and $\bm{J} \in \mathbb{R}^{6\times n}$ is the whole-body Jacobian matrix of the MM. Specifically, $\dot{\bm{q}} =[\dot{\bm{q}}_b^T \quad \dot{\bm{q}}_a^T]^T$ and $n=n_b+n_a$, where $\dot{\bm{q}}_b = \dot{\bm{x}}_b \in \mathbb{R}^{n_b}$ and $\dot{\bm{q}}_a \in \mathbb{R}^{n_a}$ are the joint velocity of mobile base and robotics arm, respectively.  

\subsubsection{Learned Mobile Base Pose Tracking} 
the goal is to track the learned mobile base trajectories for the secondary task without violating the EE's desired trajectories. Therefore, the task is defined as:
\begin{equation}
    \min_{\dot{\bm{q}}} \,\| \bm{H}_b \bm{q} -\bm{q}_{b,d} \|^2 + \| \bm{H}_a\bm{q} - \bm{q}_{a,mid} \|^2,
\end{equation}
which is written as a function of the optimization variable:
\begin{equation} \label{eq:base_track}
    \min_{\dot{\bm{q}}} \,\| \bm{H}_b\dot{\bm{q}}\Delta t-(\bm{q}_{b,d}-\bm{q}_{b}^{*}) \|^2 + \| \bm{H}_a\dot{\bm{q}}\Delta t-(\bm{q}_{a,mid}-\bm{q}_{a}^{*}) \|^2,
\end{equation}
where $\bm{H}_a, \bm{H}_b \in \mathbb{R}^n$ are selection matrices. $\Delta t \in \mathbb{R}^+$ is the control period, and $\bm{q}_{b}^{*} \in \mathbb{R}^{n_b},\bm{q}_{a}^{*} \in \mathbb{R}^{n_a}$ are the integrated optimal solutions obtained by the HQP at the previous time instant, where $\bm{q}^{*} = [ {\bm{q}_{b}^{*}}^T\quad {\bm{q}_{a}^{*}}^T ]^T $.
Besides, $(\bm{q}_{a,mid}=(\bm{q}_{a,min}+\bm{q}_{a,max})/2) \in \mathbb{R}^{n_a}$, where $\bm{q}_{a,min}, \bm{q}_{a,max}\in \mathbb{R}^{n_a}$ are the minimum and maximum mechanical limits of the robotic arm joints.
The first term in \eqref{eq:base_track} tracks the learned mobile base pose, and the second term keeps the robotic arm away from joint limits.

\subsubsection{Constraints}
the constraints that regulate the HQP problem are defined at position, velocity, and acceleration levels based on the physical limits of the MM's actuators.
By leaving the floating base Cartesian coordinates unbounded, only the arm constraints are imposed at position level
\begin{equation} \label{eq:arm_constraints}
\bm{q}_{a,min} \leq \bm{q}_{a}^* +\dot{\bm{q}}_a \ \Delta t \leq \bm{q}_{a,max}.
\end{equation}
Besides, the velocity and acceleration constraints for the whole-body MM can be written as: 
\begin{gather}
        \dot{\bm{q}}_{min} \leq \dot{\bm{q}} \leq \dot{\bm{q}}_{max} \\
        \ddot{\bm{q}}_{min} \leq \frac{\dot{\bm{q}} -\dot{\bm{q}}^*}{\Delta t} \leq \ddot{\bm{q}}_{max} ,
\end{gather}
where $ \dot{\bm{q}}^* \in \mathbb{R}^n $ is the optimal joint velocity identified at the previous time instant.  
$\dot{\bm{q}}_{min}, \ddot{\bm{q}}_{min} \in \mathbb{R}^n$ and $ \dot{\bm{q}}_{max}, \ddot{\bm{q}}_{max} \in \mathbb{R}^n$ are the velocity and acceleration limits, respectively.

\subsection{Joint Impedance Control for Robotic Arm}\label{sec:joint_impedance_control}
To guarantee a low-level safety in the interaction, we exploit the torque-controlled arm via a decentralized joint impedance controller, which ensures a degree of compliance based on:
\begin{equation} \label{eq:joint_imp_ctrl}
    \bm{\tau}_a = \bm{K_{q_{a,d}}} (\dot{\bm{q}}^*_a - \dot{\bm{q}}_{a,act}) + \bm{K_{q_{a,p}}} (\bm{q}^*_a - \bm{q}_{a,act}) + \bm{g}_a
\end{equation}
where $\bm{q}^*_a, \dot{\bm{q}}^*_a \in \mathbb{R}^{n_a}$ represent the desired robotic arm optimal joints' position and velocity, respectively, obtained from the HQP. Concurrently, $\bm{q}_{a,act}, \dot{\bm{q}}_{a,act} \in \mathbb{R}^{n_a}$ denote the actual joint positions and velocities. The matrices $\bm{K_{q_{a,p}}} \in \mathbb{R}^{n_a\times n_a}$ and $\bm{K_{q_{a,d}}} \in \mathbb{R}^{n_a\times n_a}$ are positive definite joint stiffness and damping matrices, respectively, and $\bm{g_a} \in \mathbb{R}^{n_a}$ corresponds to the gravity compensation vector.

Notably, the optimal joint velocity $\dot{\bm{q}}^*_a$ from the HQP could be directly applied to a default joint velocity controller for a robotic arm that does not support joint torque control.
However, this added layer guarantees safety in case of unforeseen contacts and impacts along any point of the kinematic chain at the expense of deviating from the desired behavior imposed at the EE via \eqref{eq:min_kyn_clik}. 
As for the mobile base, the optimal joint velocity $\dot{\bm{q}}^*_b\in\mathbb{R}^{n_b}$ is sent directly to the default controller since most of the mobile platforms only provide velocity control interfaces.

\section{Experiments and Results}\label{sec:exp}
We used the MObile Collaborative robotic Assistant (MOCA)~\cite{Zhao2022} platform to evaluate the proposed approach in a locomotion-integrated pick-and-place task. MOCA is a mobile manipulator with a joint torque-controlled 7-DoFs Franka Emika Panda robotic arm mounted on top of the velocity-controlled 3-DoFs Robotnik SUMMIT-XL STEEL mobile base. Firstly, human demonstrations were collected (see Fig. \ref{fig:exp_setup}), where the subject picked a bottle without stopping and placed it at the desired point later while walking. The demonstrations were learned and adapted to the new initial positions. After that, MOCA was employed to finish the learned and adapted task in the real world. Besides, the Franka default gripper with custom fingers was used to execute the task, and the overall robotic platform is in Fig. \ref{fig:snapshots}. The details are described in the following sections.   

\subsection{Human Demonstrations Collection}\label{subsec:human-demo}
\begin{figure}[t]
    \centering 
    \includegraphics[trim=2.0cm 0.0cm 0.0cm 0.0cm,clip,width=\columnwidth]{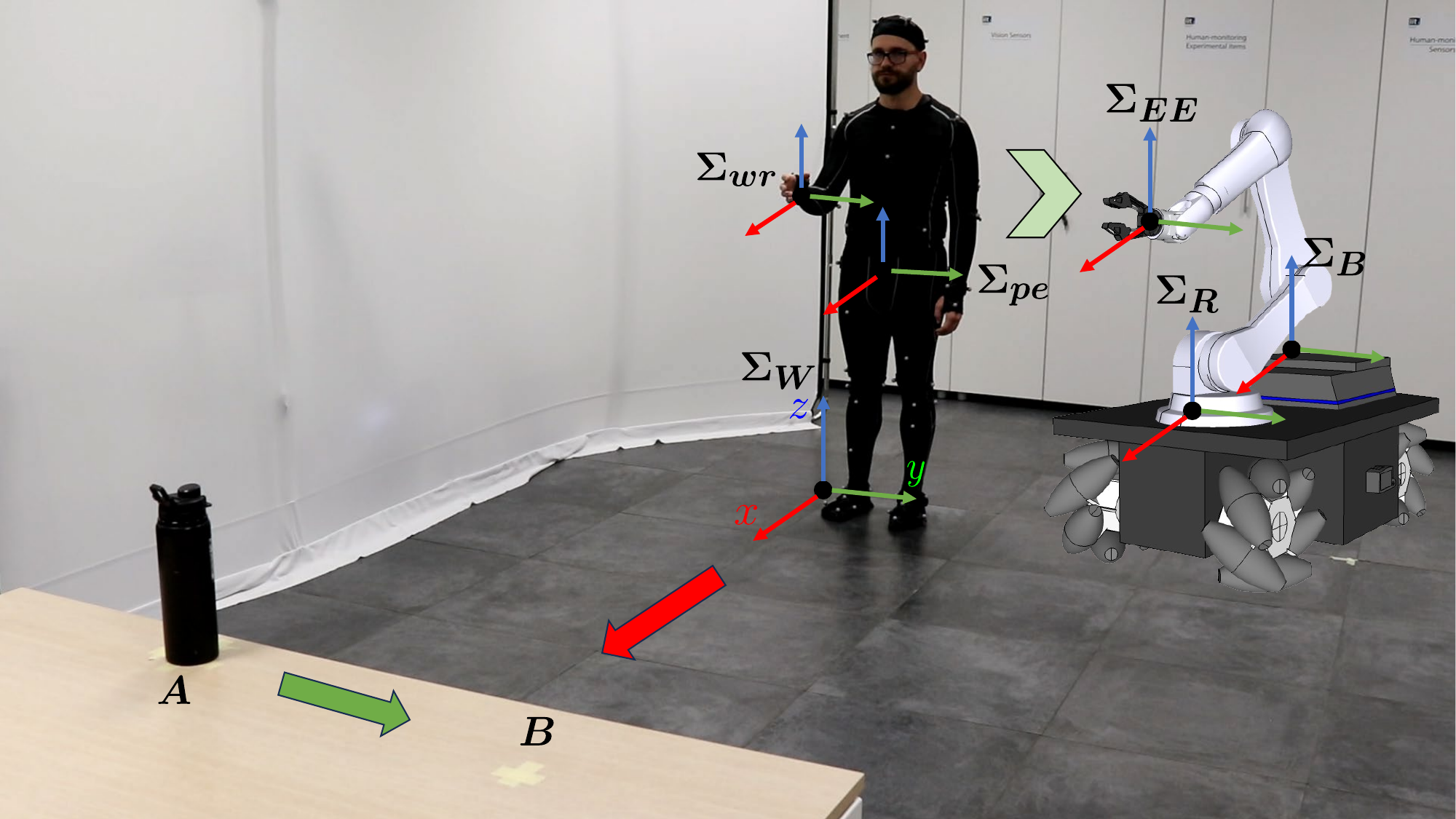} 
    \caption{Experimental setting for human demonstration. The subject was asked to naturally pick the bottle at point $A$ while walking at a natural speed and to place it at point $B$. All the motions were described in the world frame $\bm{\Sigma_{W}}$. The frames of the subject and MOCA are illustrated.}
    \label{fig:exp_setup}
\end{figure}

A healthy human subject was asked to perform the pick-and-place task to collect demonstrations and analyze human whole-body behavior. The overall setup is shown in Fig.~\ref{fig:exp_setup}.

\subsubsection{Setup}

The subject was asked to wear a Lycra suit equipped with markers tracked in real-time by the OptiTrack system to capture comprehensive data on the whole-body motion. The integration and synchronization of data from various sensors were orchestrated within the Robot Operating System (ROS) environment at a sampling rate of 100 Hz.

\subsubsection{Protocol} 
\begin{figure*}[t]
    \centering
    \includegraphics[trim=0.0cm 0.0cm 3.0cm 0.0cm,clip,width=0.27\columnwidth]{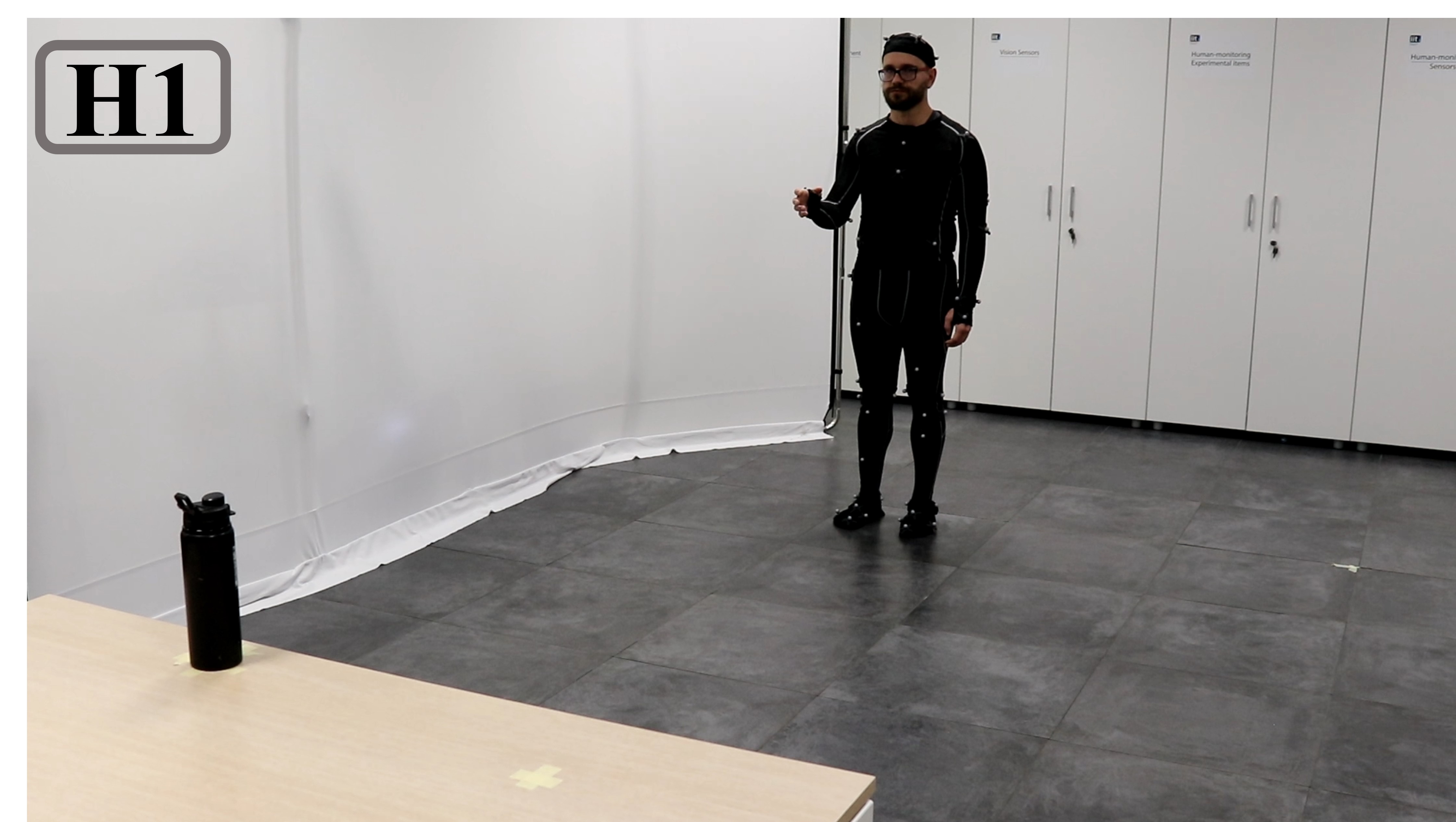}
    \includegraphics[trim=0.0cm 0.0cm 3.0cm 0.0cm,clip,width=0.27\columnwidth]{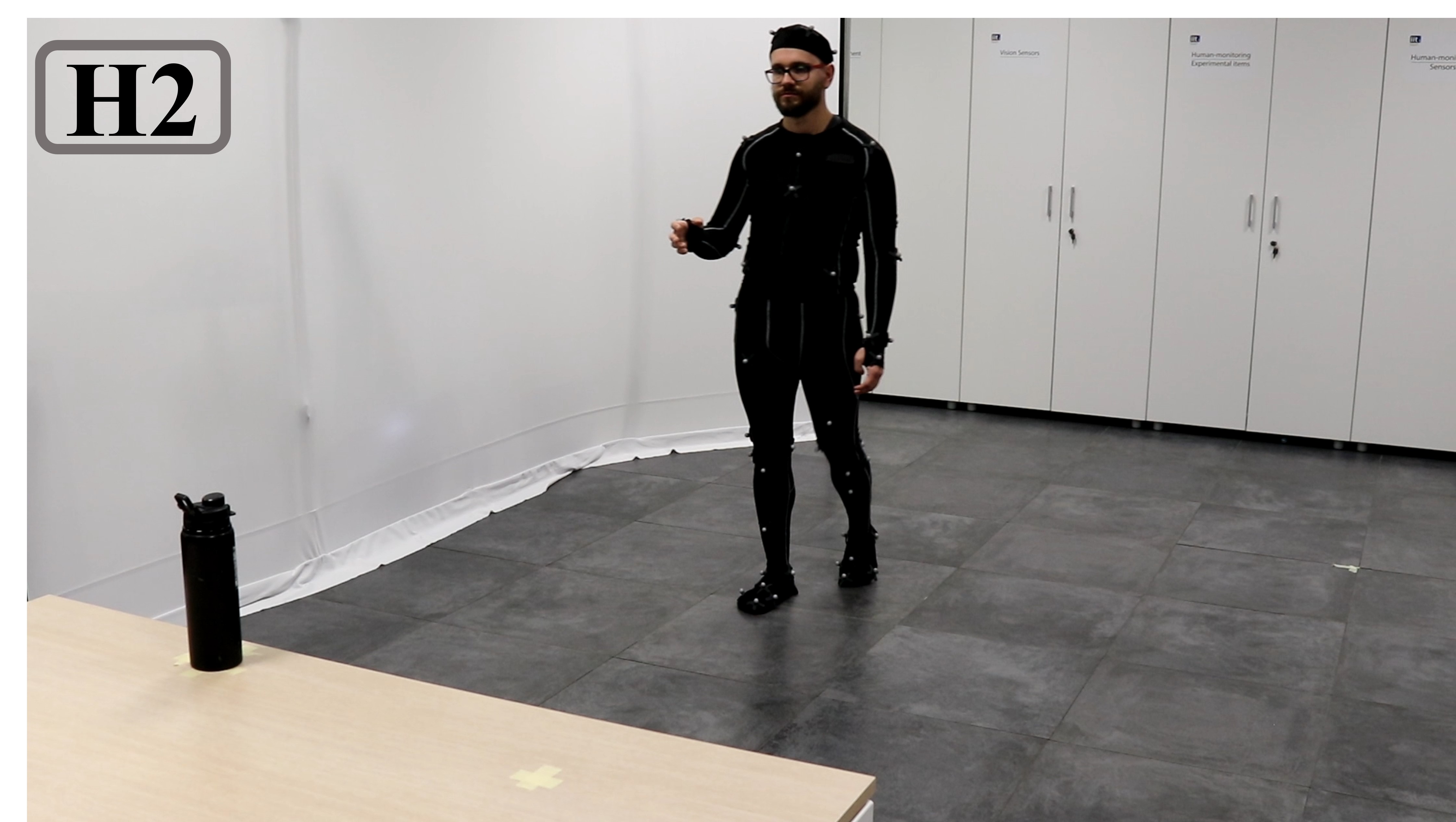}
    \includegraphics[trim=0.0cm 0.0cm 3.0cm 0.0cm,clip,width=0.27\columnwidth]{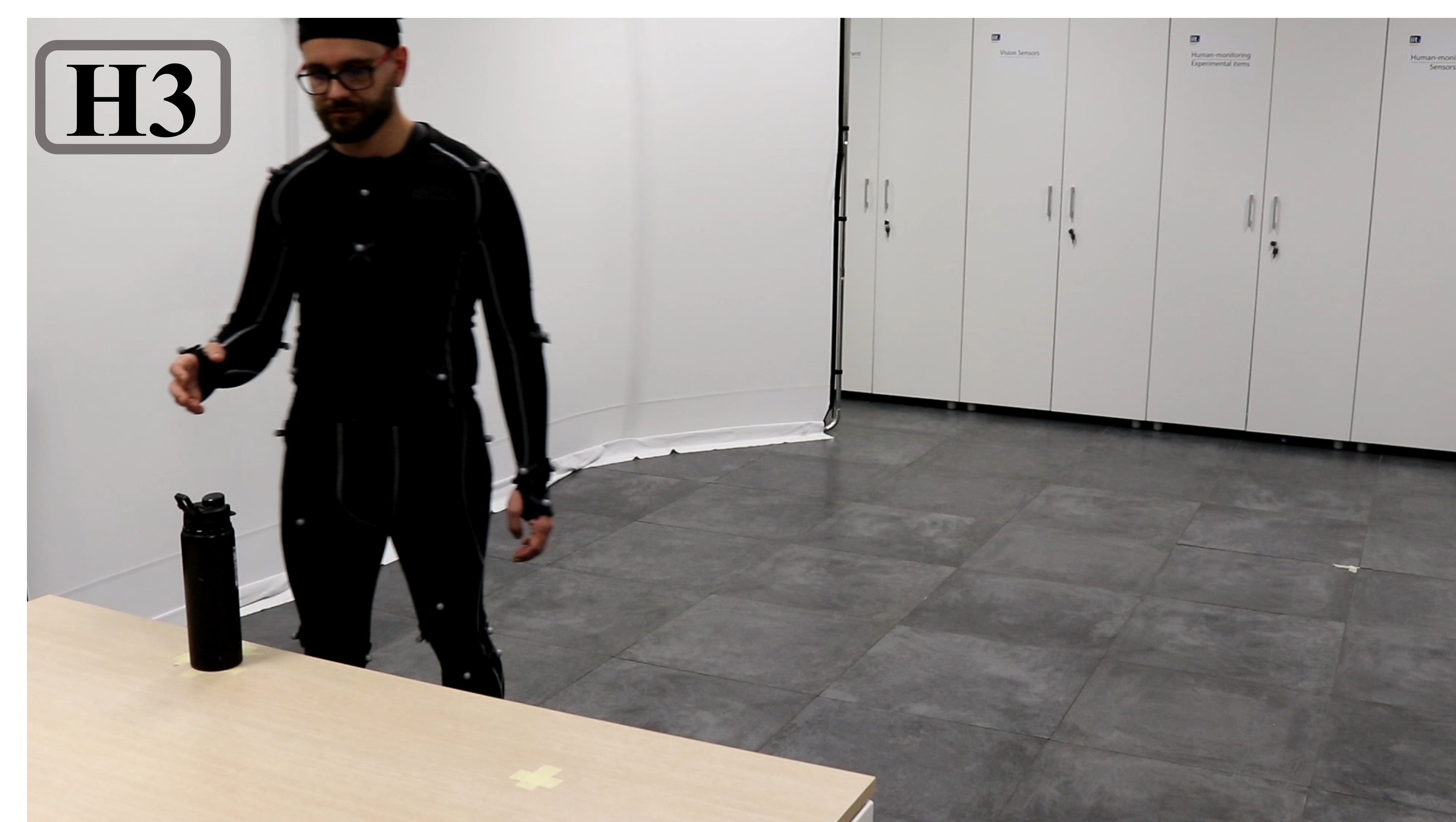}
    \includegraphics[trim=0.0cm 0.0cm 3.0cm 0.0cm,clip,width=0.27\columnwidth]{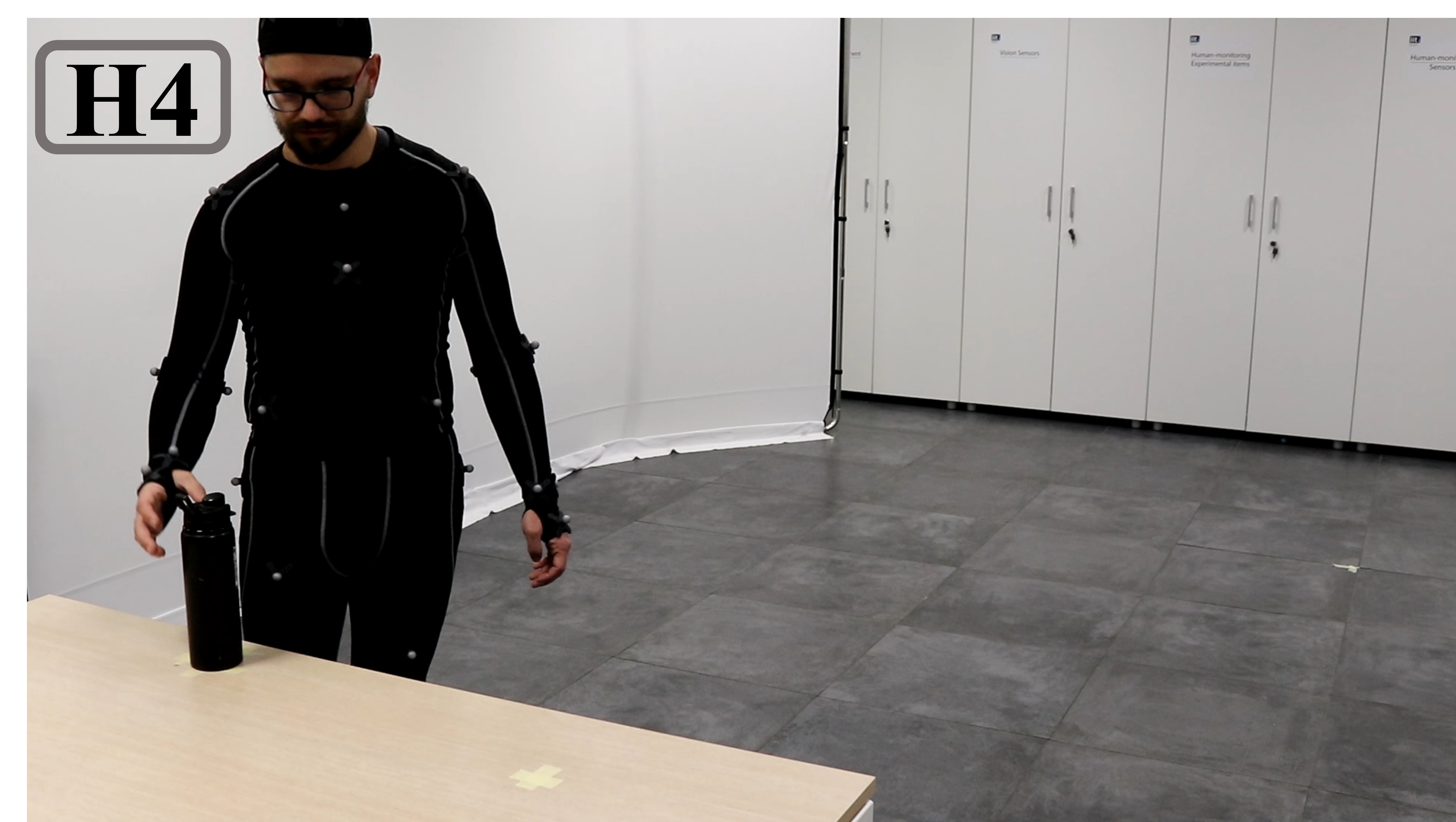}
    \includegraphics[trim=0.0cm 0.0cm 3.0cm 0.0cm,clip,width=0.27\columnwidth]{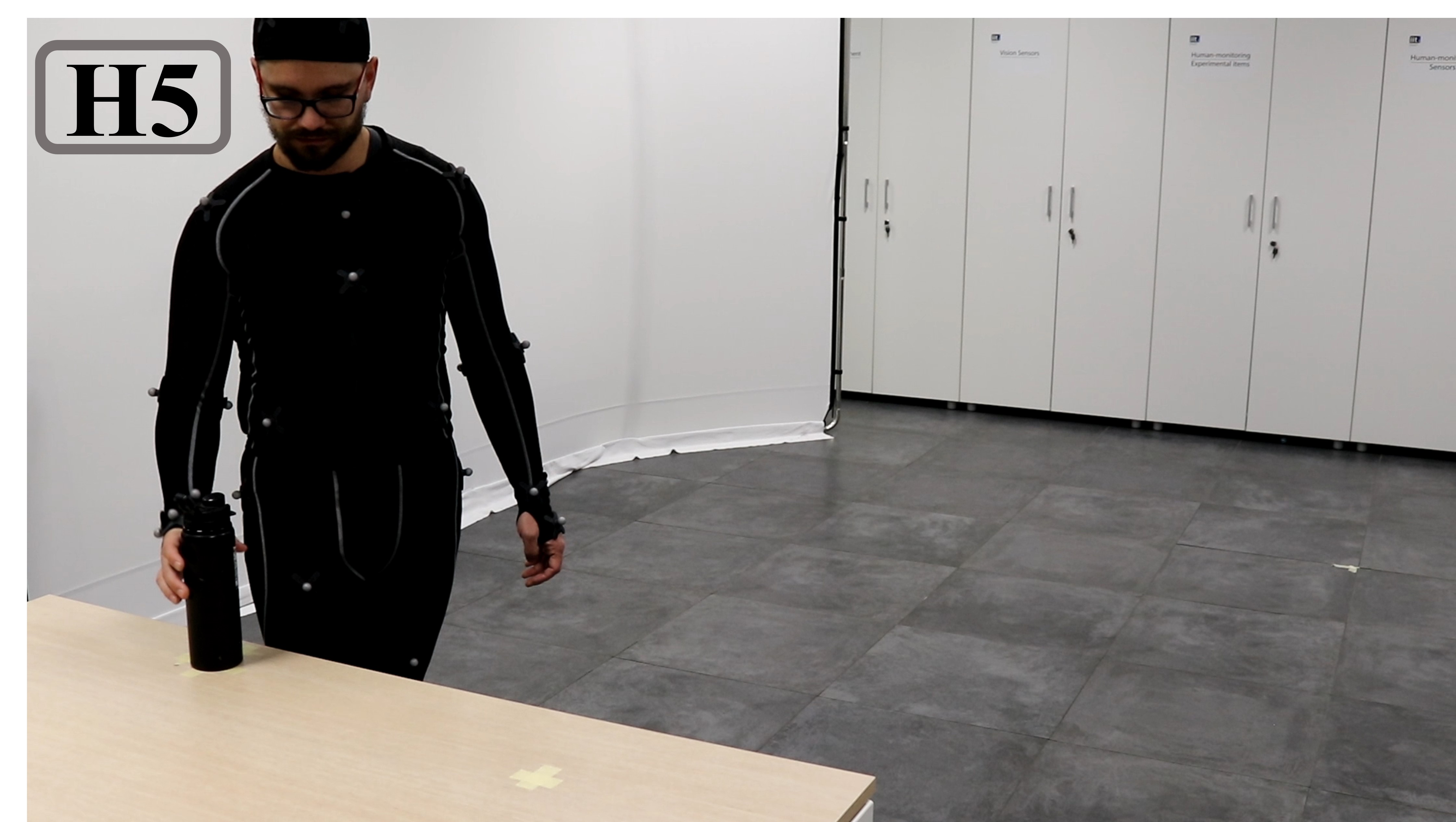}
    \includegraphics[trim=0.0cm 0.0cm 3.0cm 0.0cm,clip,width=0.27\columnwidth]{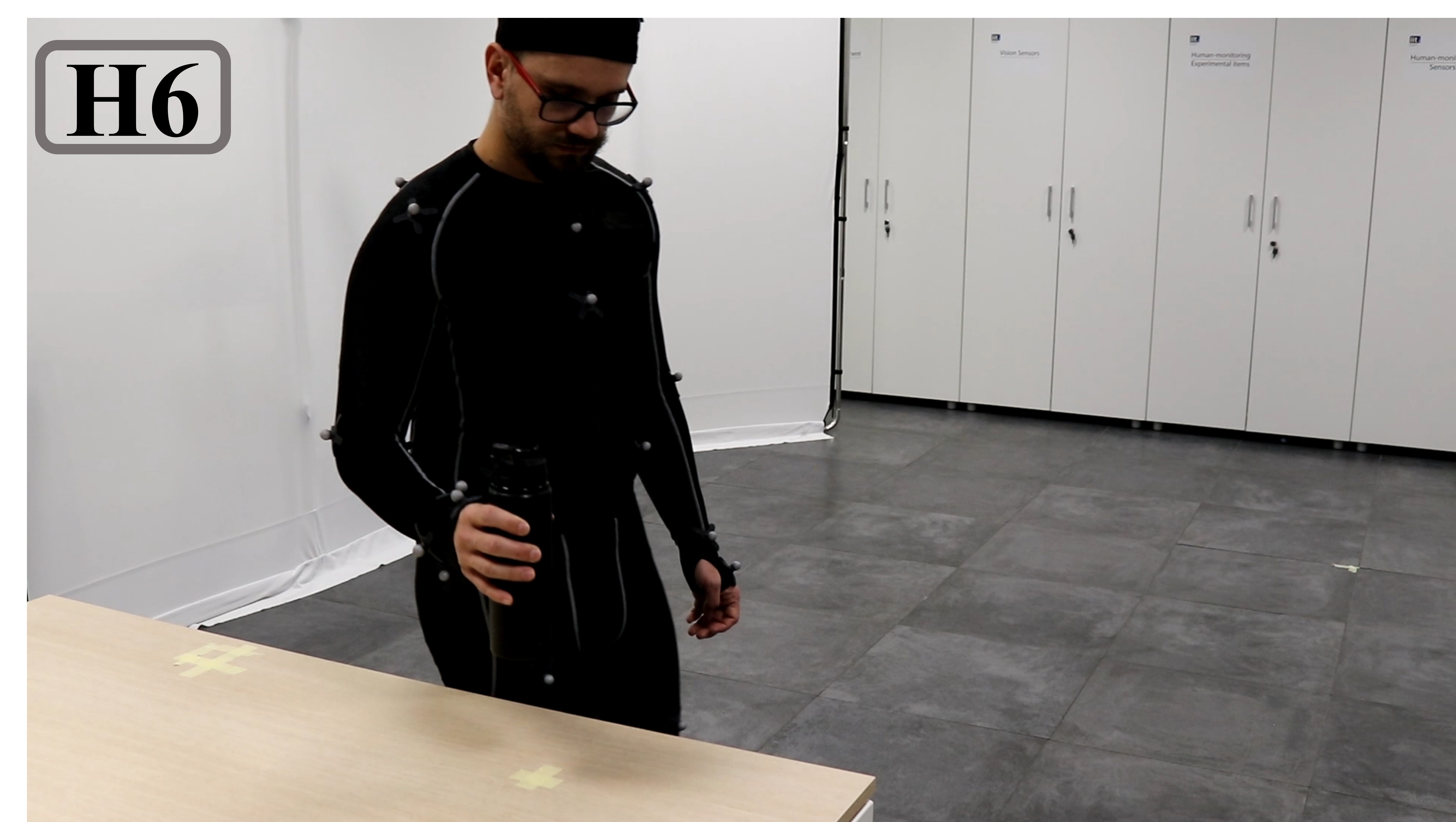}
    \includegraphics[trim=0.0cm 0.0cm 3.0cm 0.0cm,clip,width=0.27\columnwidth]{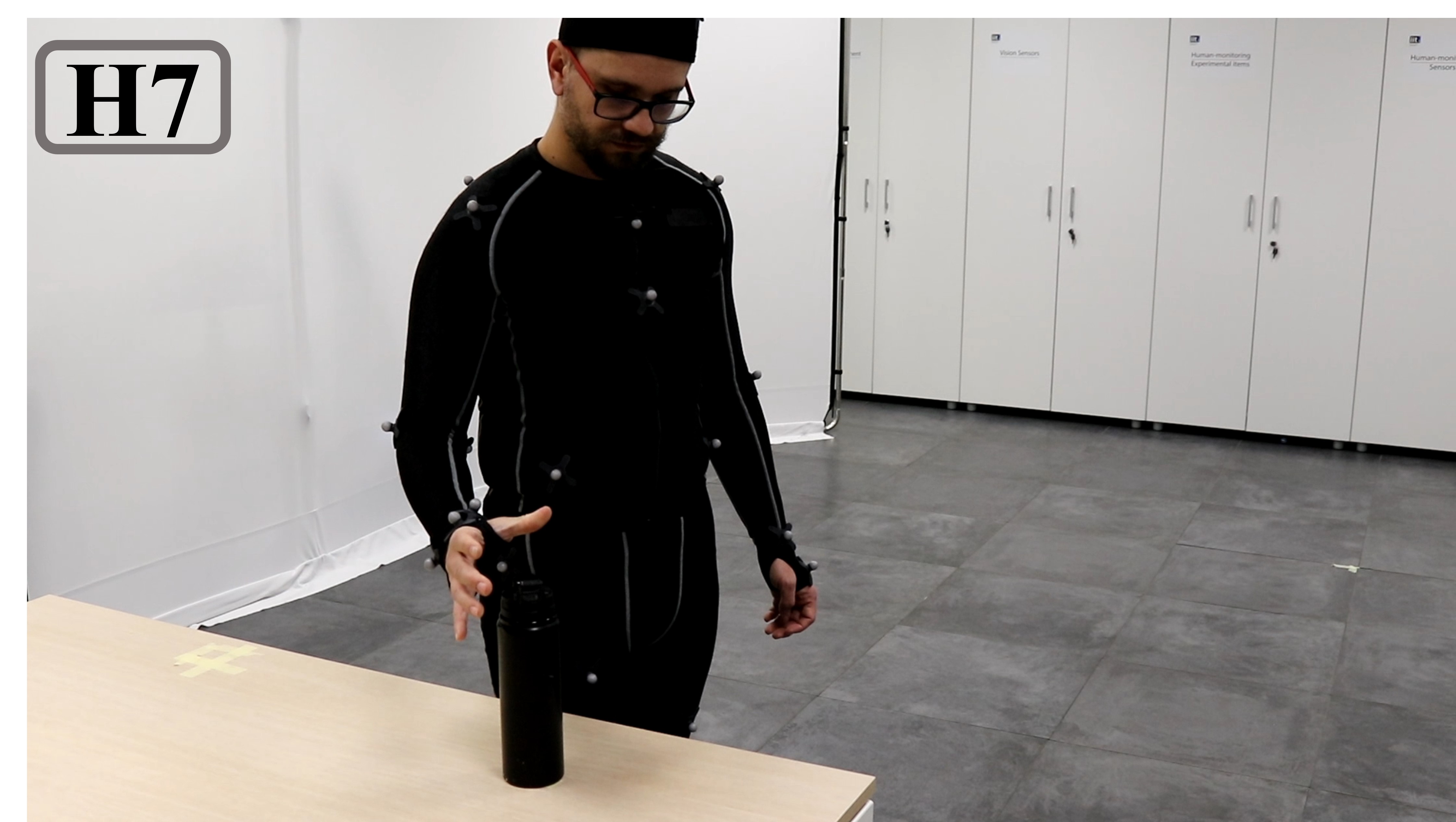}\\
    \vspace{2mm}
    \includegraphics[trim=0.0cm 0.0cm 3.0cm 0.0cm,clip,width=0.27\columnwidth]{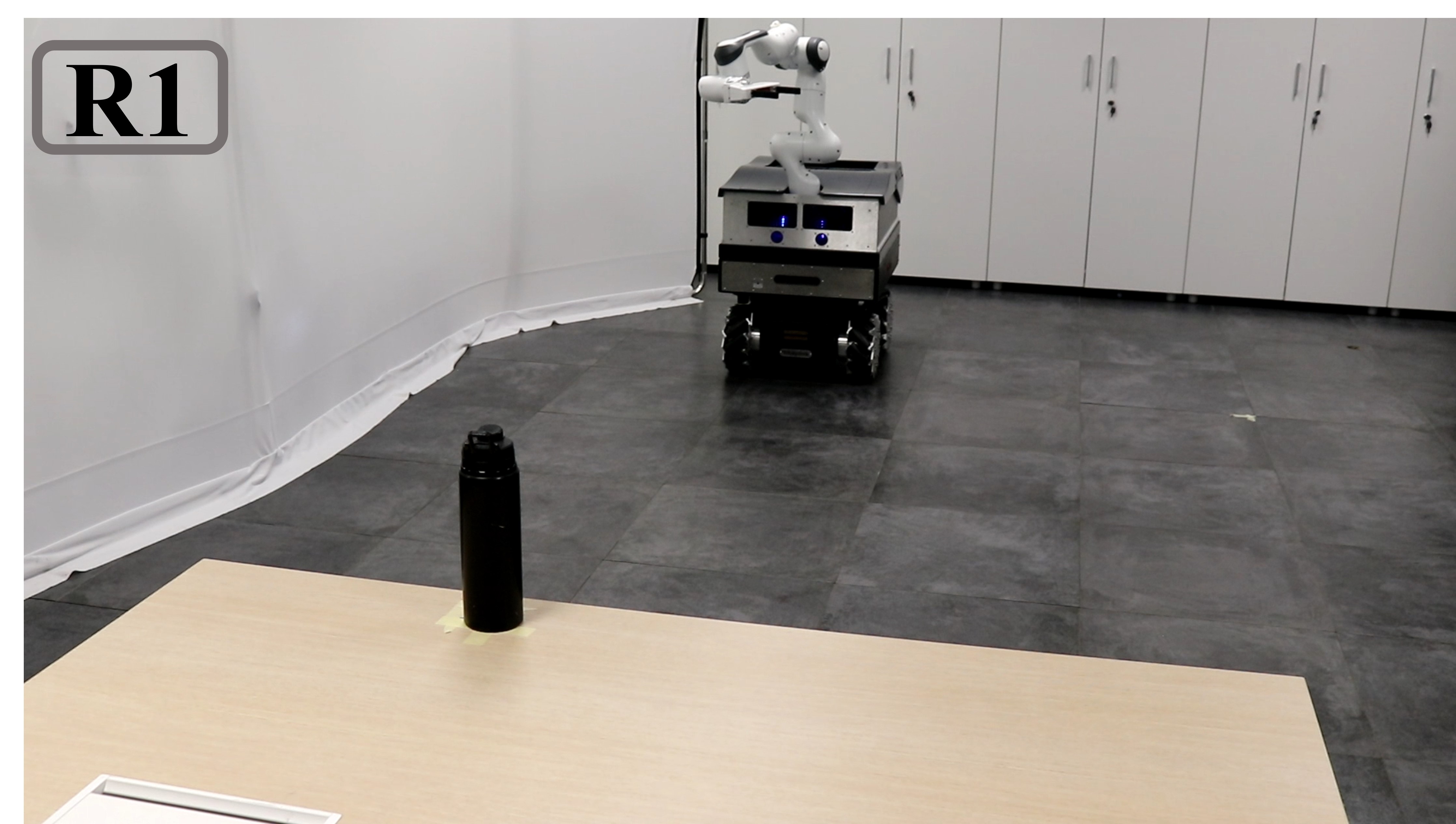}
    \includegraphics[trim=0.0cm 0.0cm 3.0cm 0.0cm,clip,width=0.27\columnwidth]{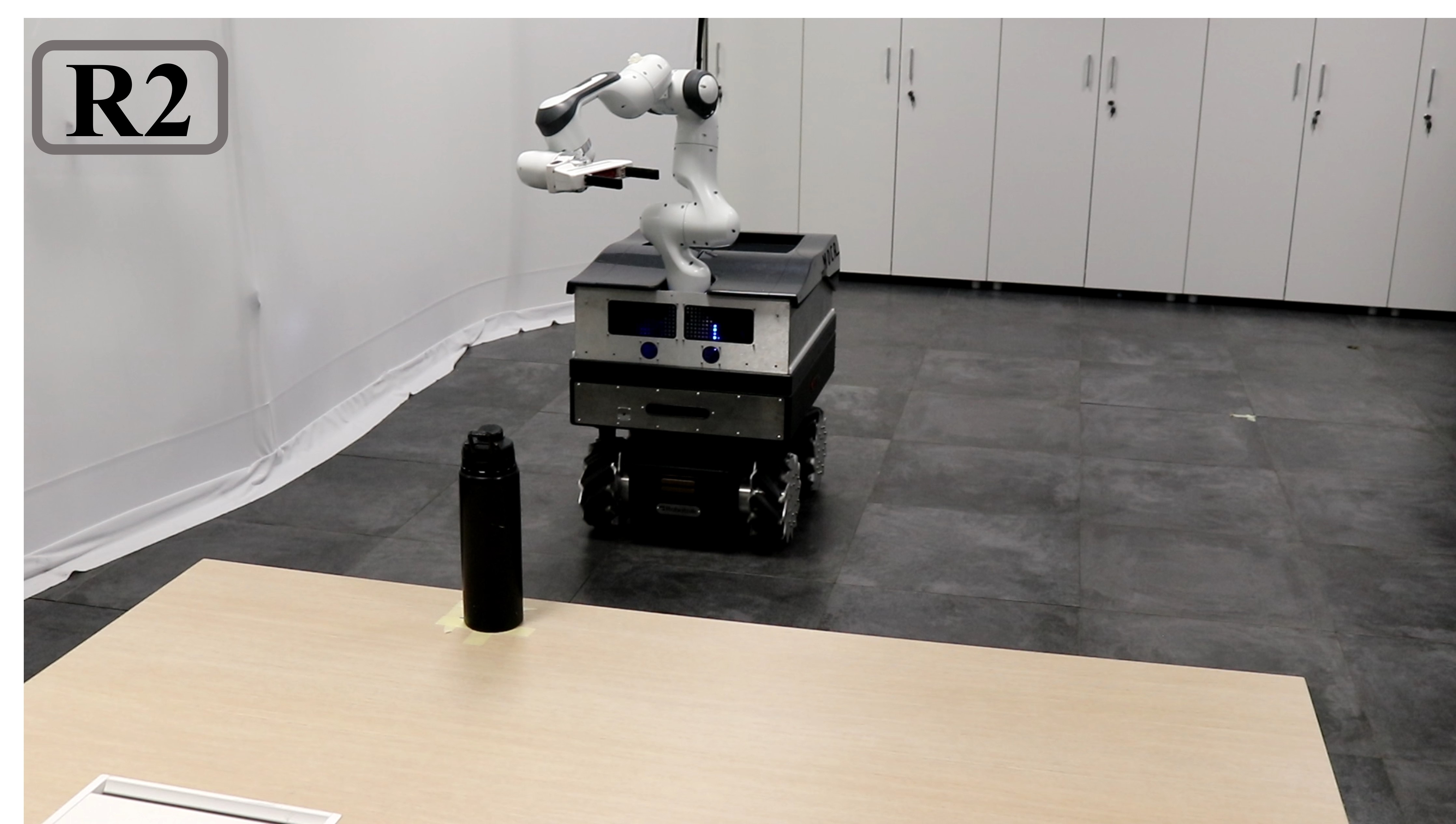}
    \includegraphics[trim=0.0cm 0.0cm 3.0cm 0.0cm,clip,width=0.27\columnwidth]{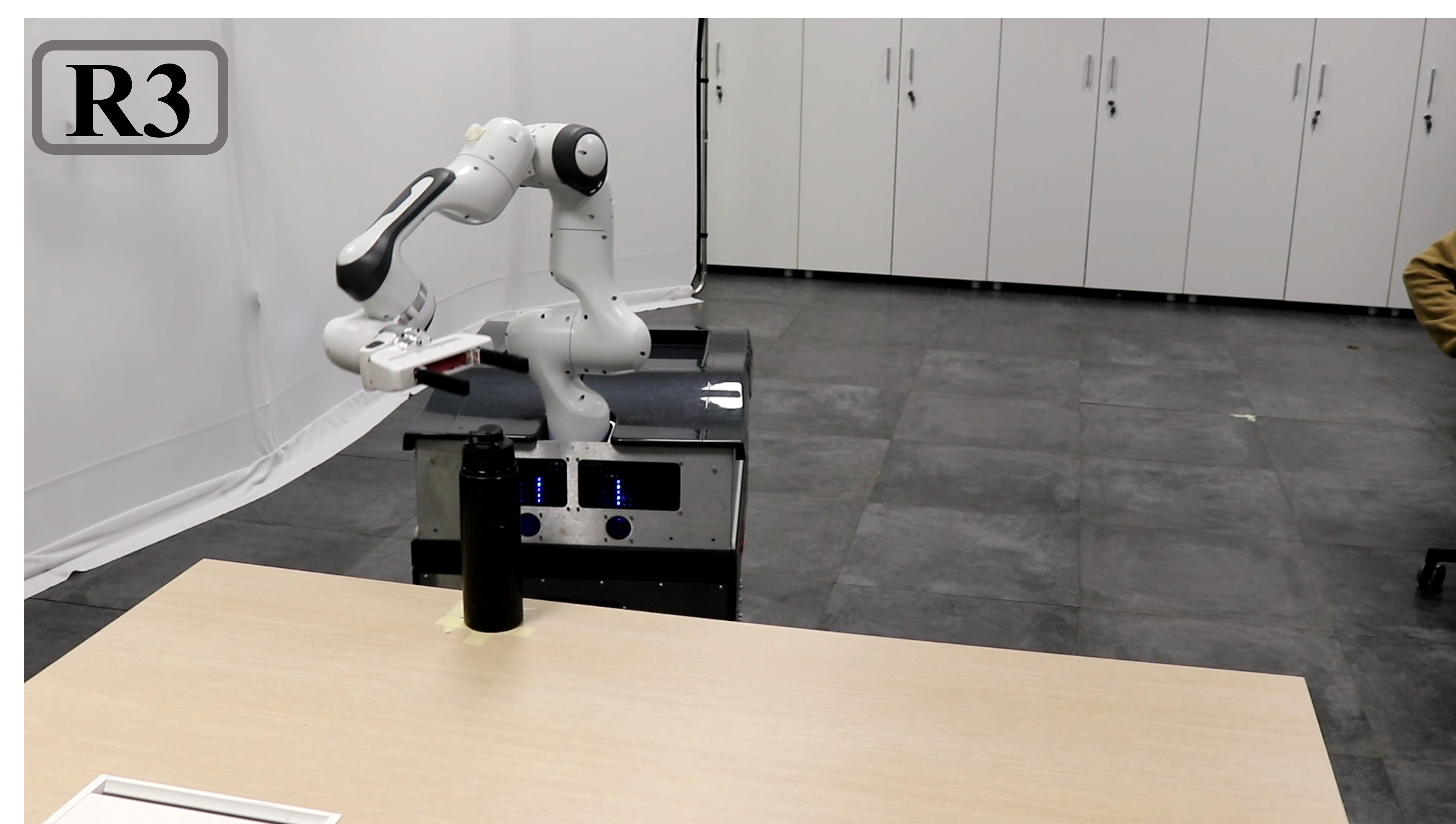}
    \includegraphics[trim=0.0cm 0.0cm 3.0cm 0.0cm,clip,width=0.27\columnwidth]{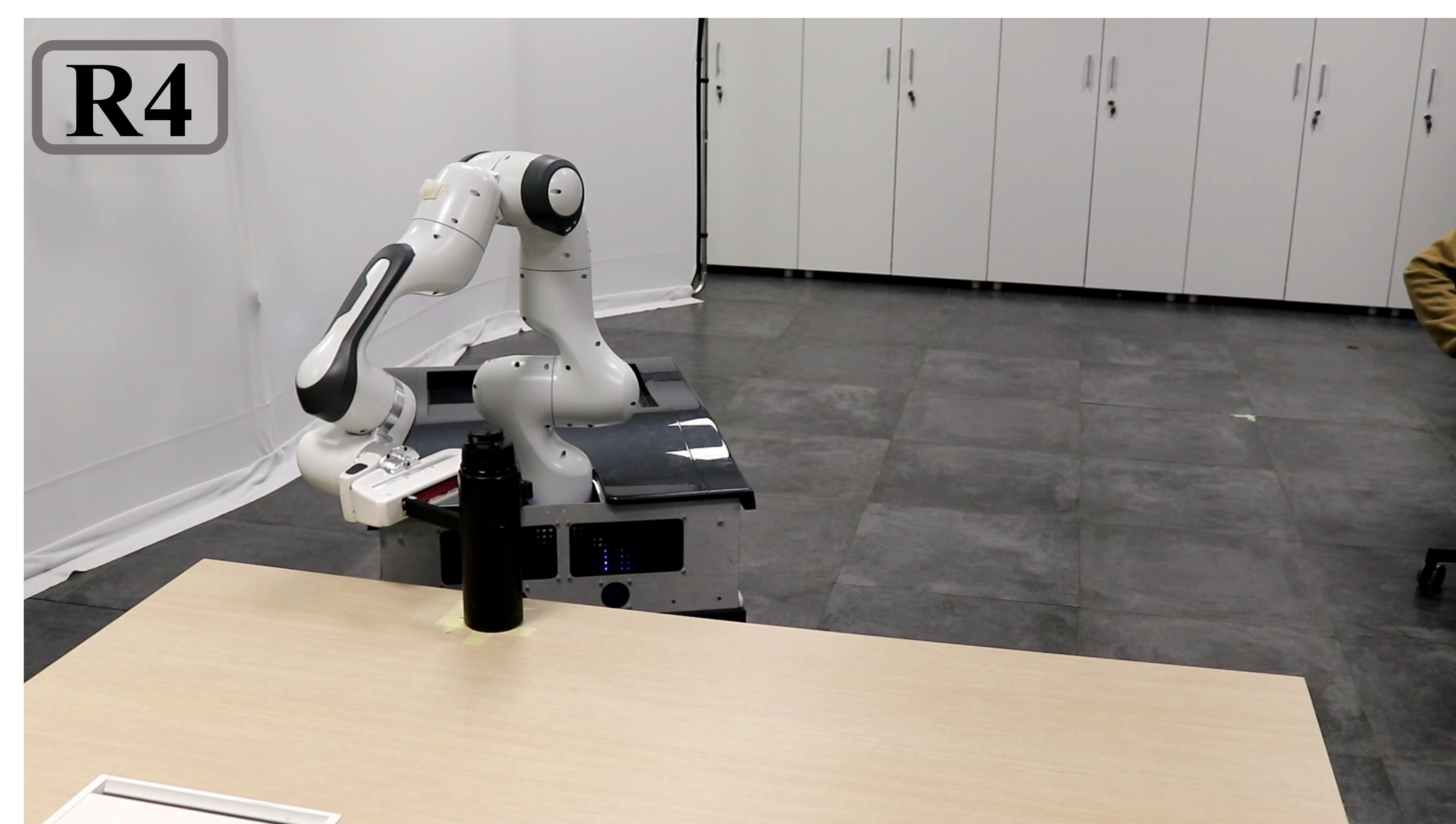}
    \includegraphics[trim=0.0cm 0.0cm 3.0cm 0.0cm,clip,width=0.27\columnwidth]{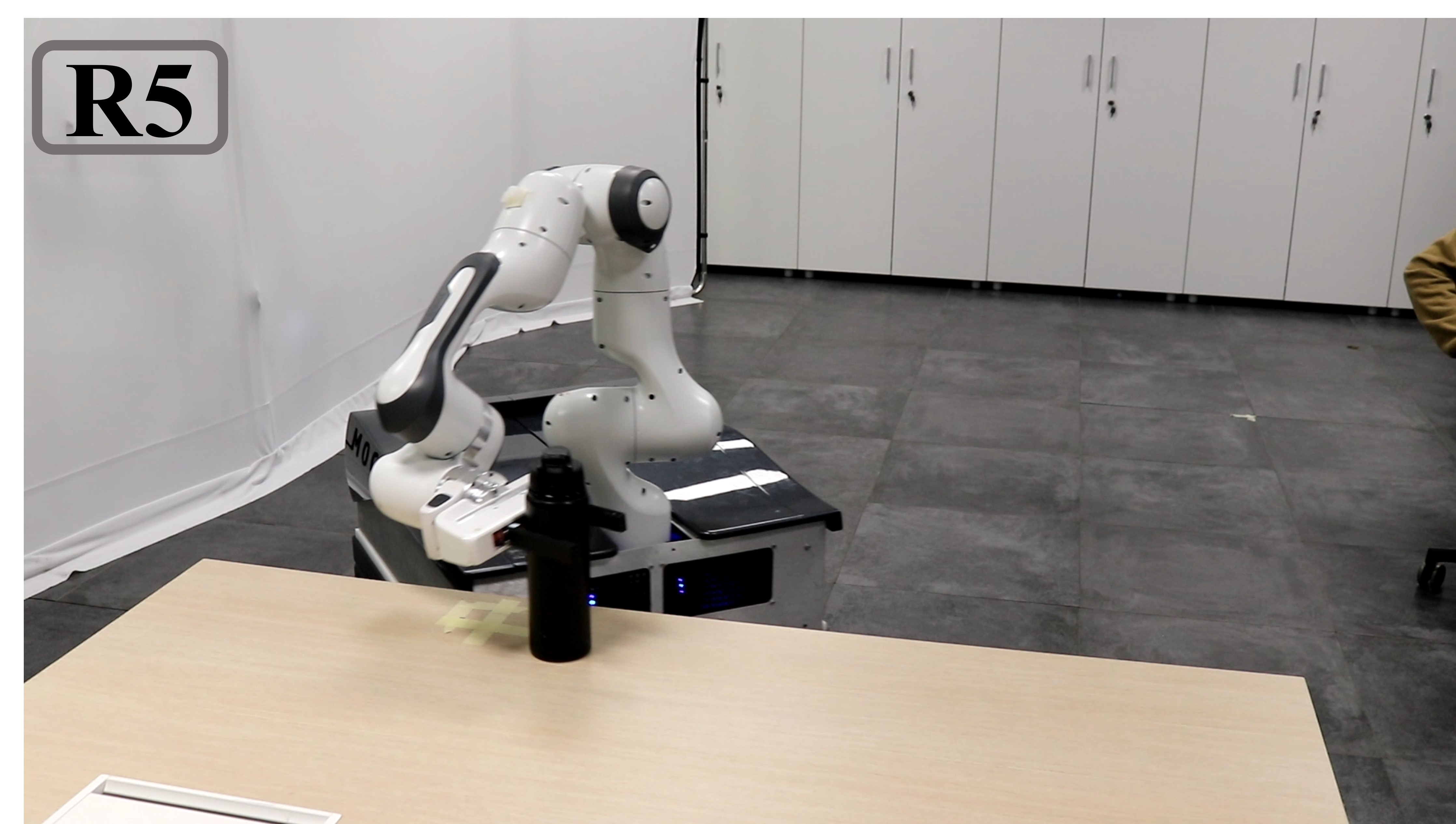}
    \includegraphics[trim=0.0cm 0.0cm 3.0cm 0.0cm,clip,width=0.27\columnwidth]{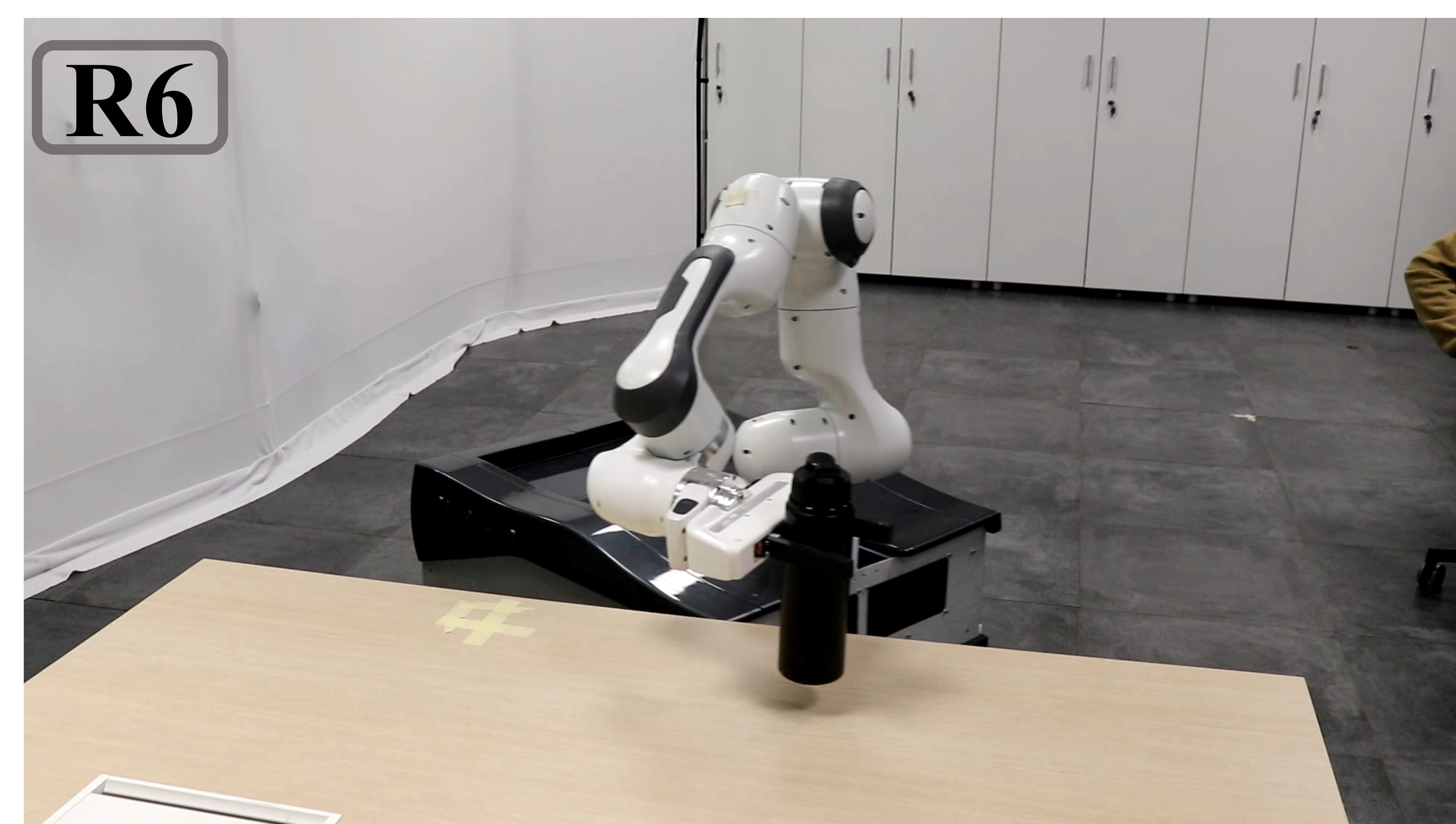}
    \includegraphics[trim=0.0cm 0.0cm 3.0cm 0.0cm,clip,width=0.27\columnwidth]{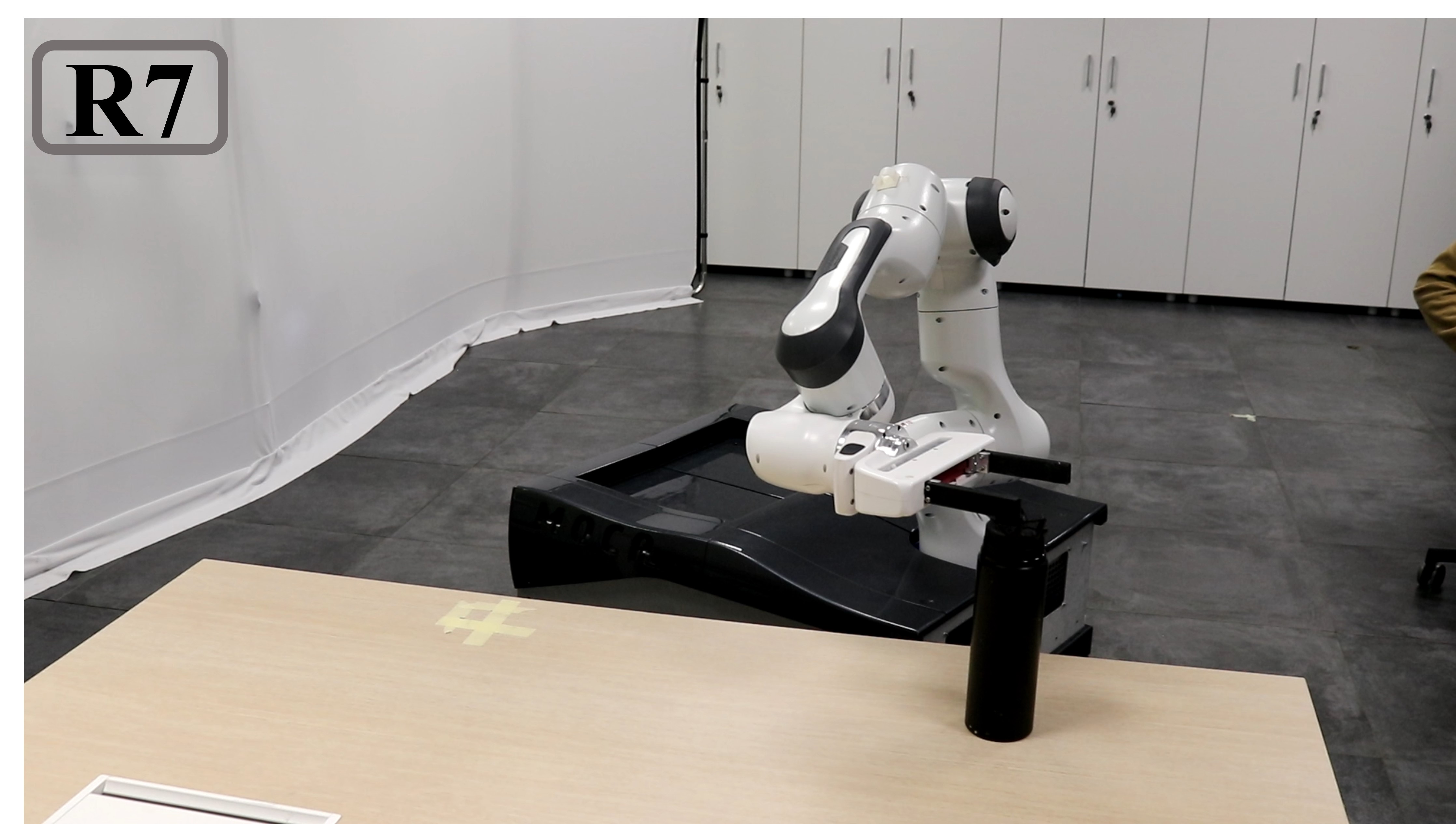}\\
    \vspace{2mm}
    \includegraphics[trim=0.0cm 0.0cm 3.0cm 0.0cm,clip,width=0.27\columnwidth]{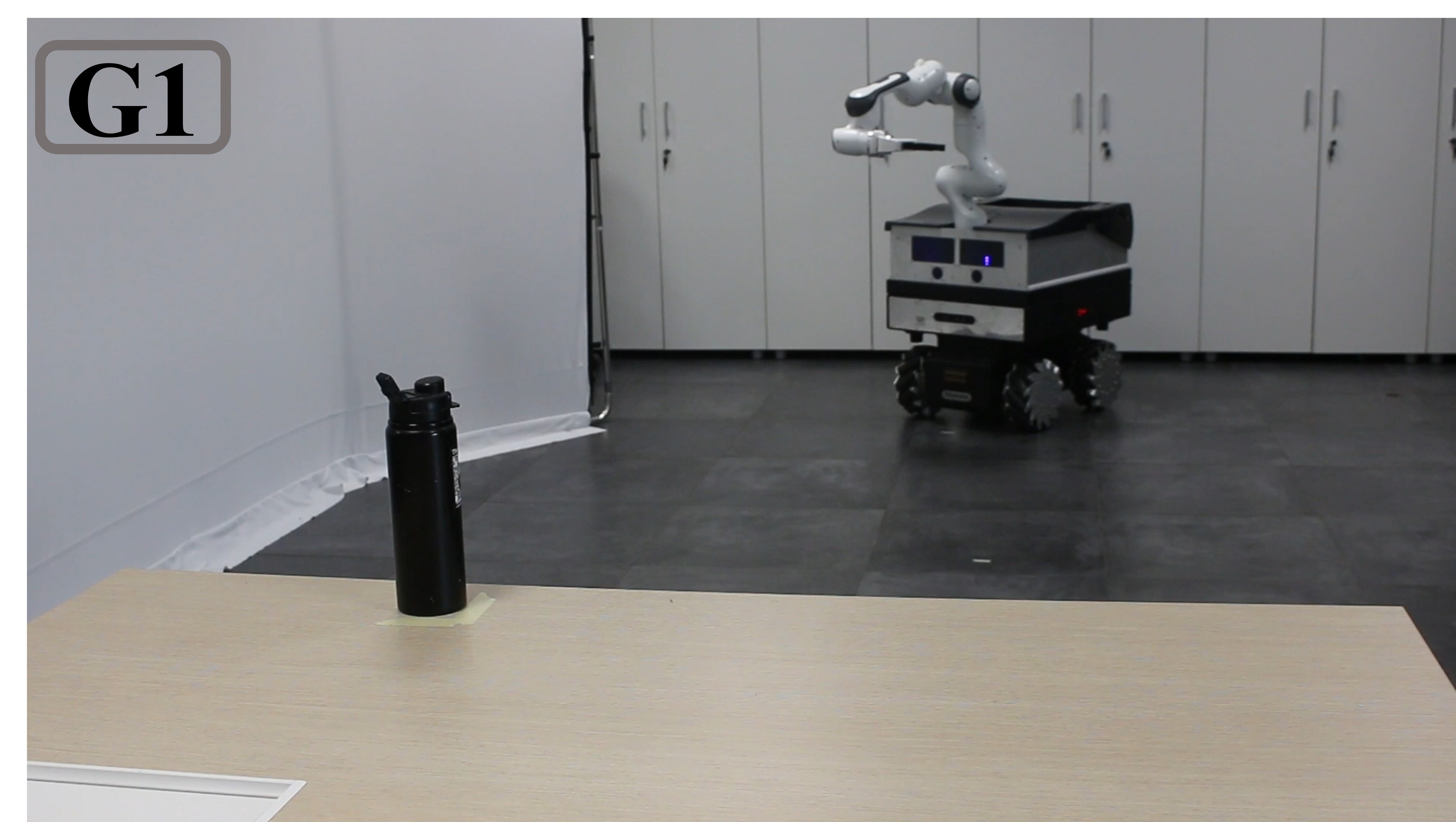}
    \includegraphics[trim=0.0cm 0.0cm 3.0cm 0.0cm,clip,width=0.27\columnwidth]{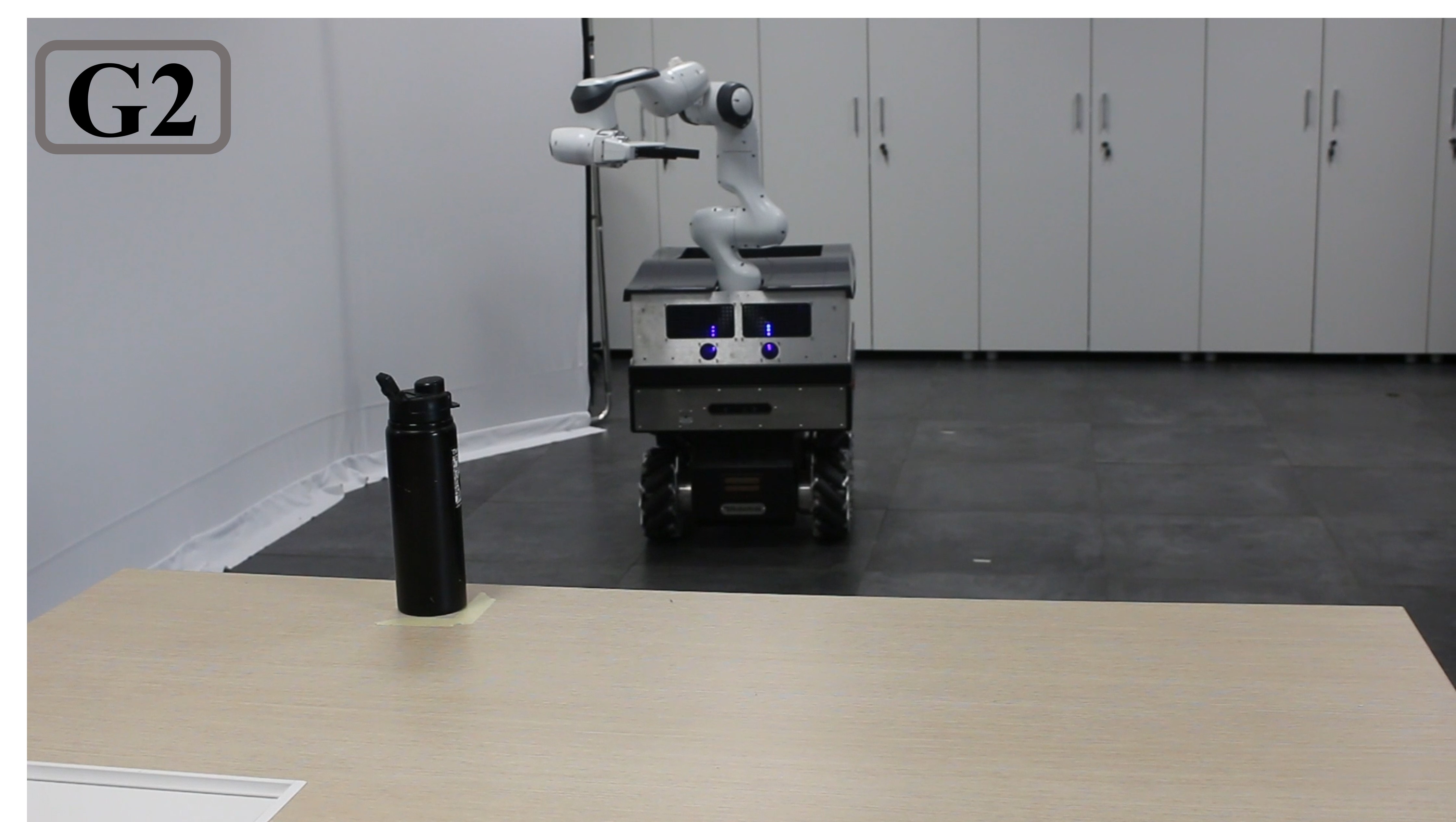}
    \includegraphics[trim=0.0cm 0.0cm 3.0cm 0.0cm,clip,width=0.27\columnwidth]{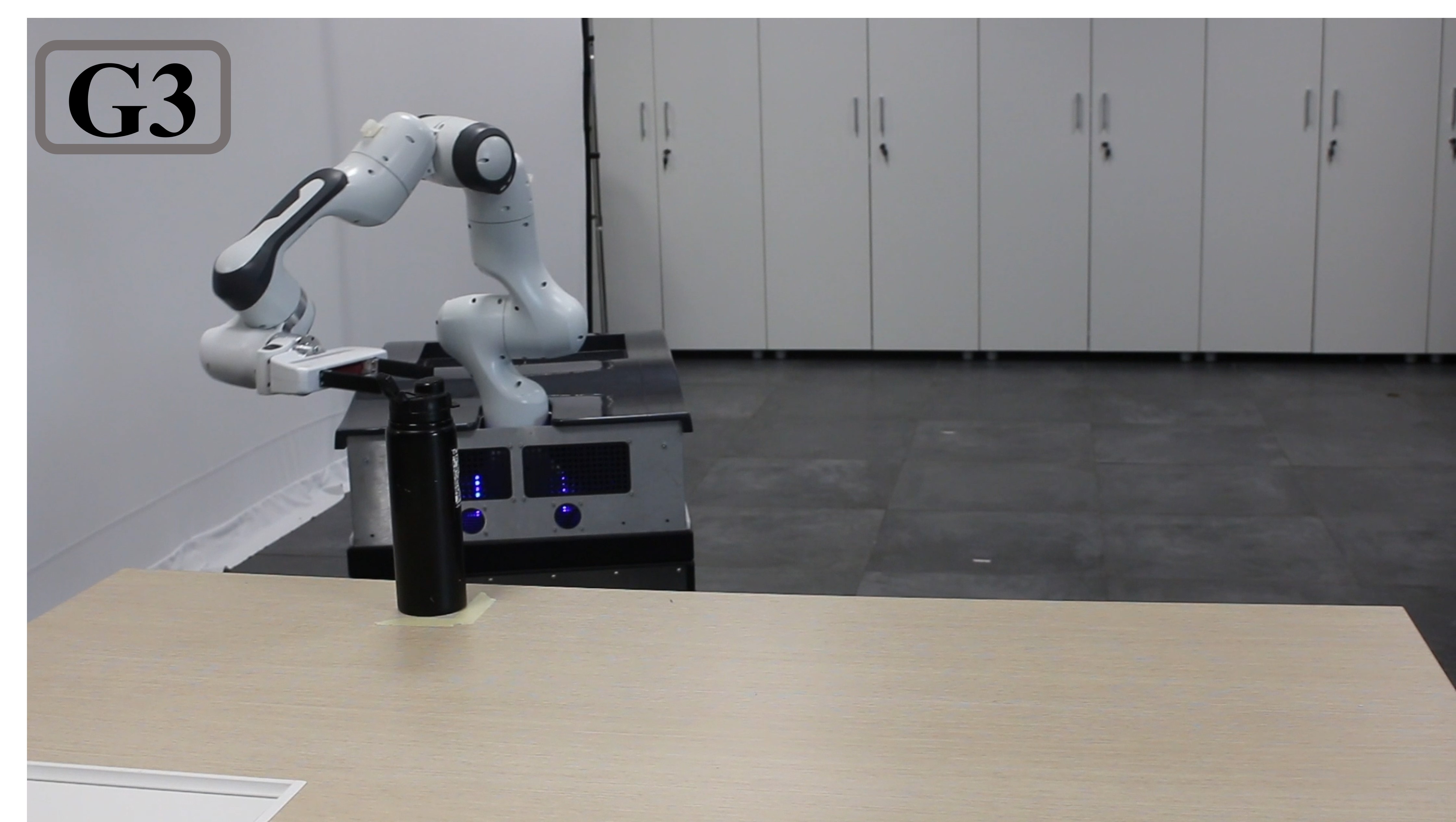}
    \includegraphics[trim=0.0cm 0.0cm 3.0cm 0.0cm,clip,width=0.27\columnwidth]{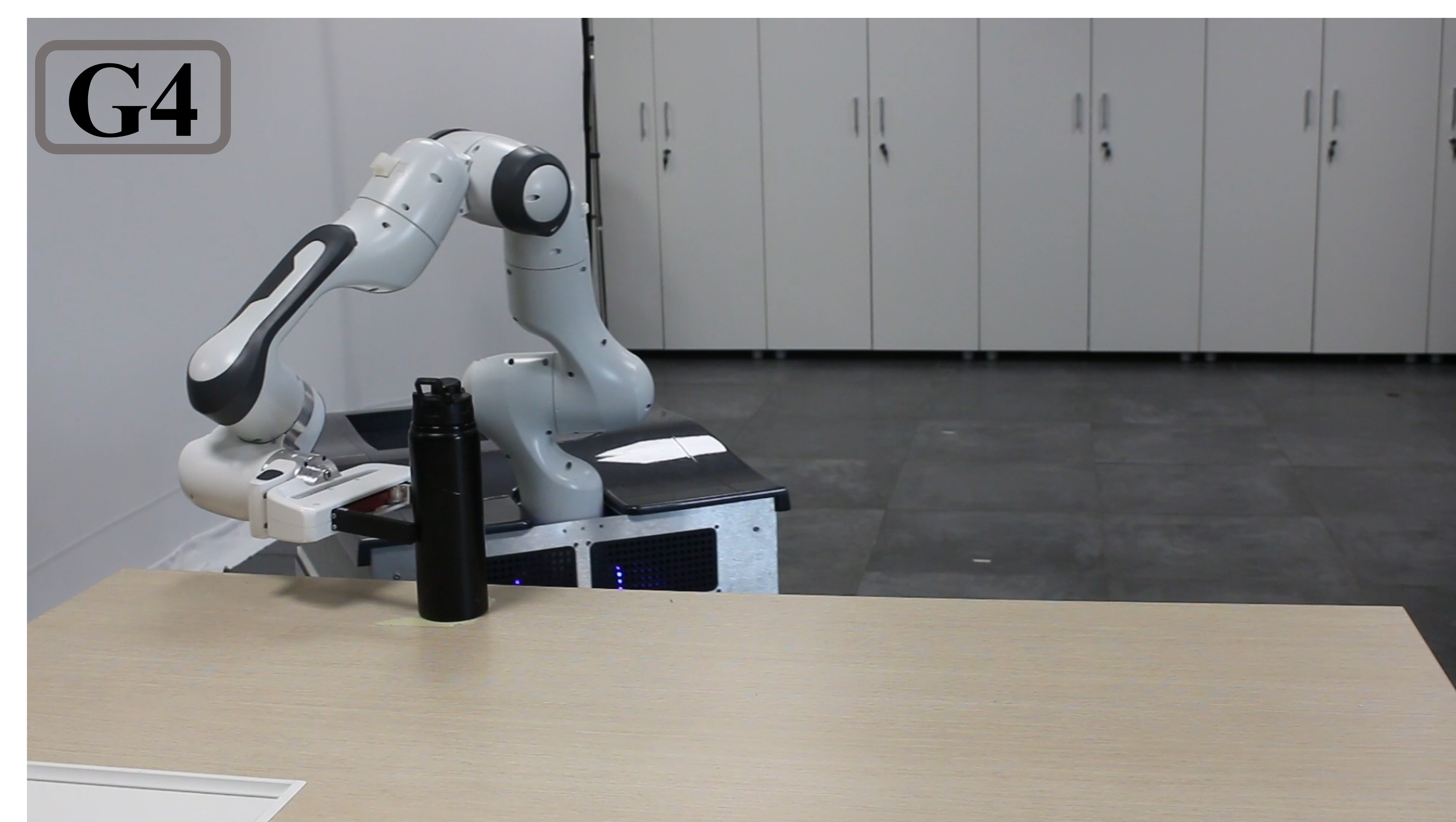}
    \includegraphics[trim=0.0cm 0.0cm 3.0cm 0.0cm,clip,width=0.27\columnwidth]{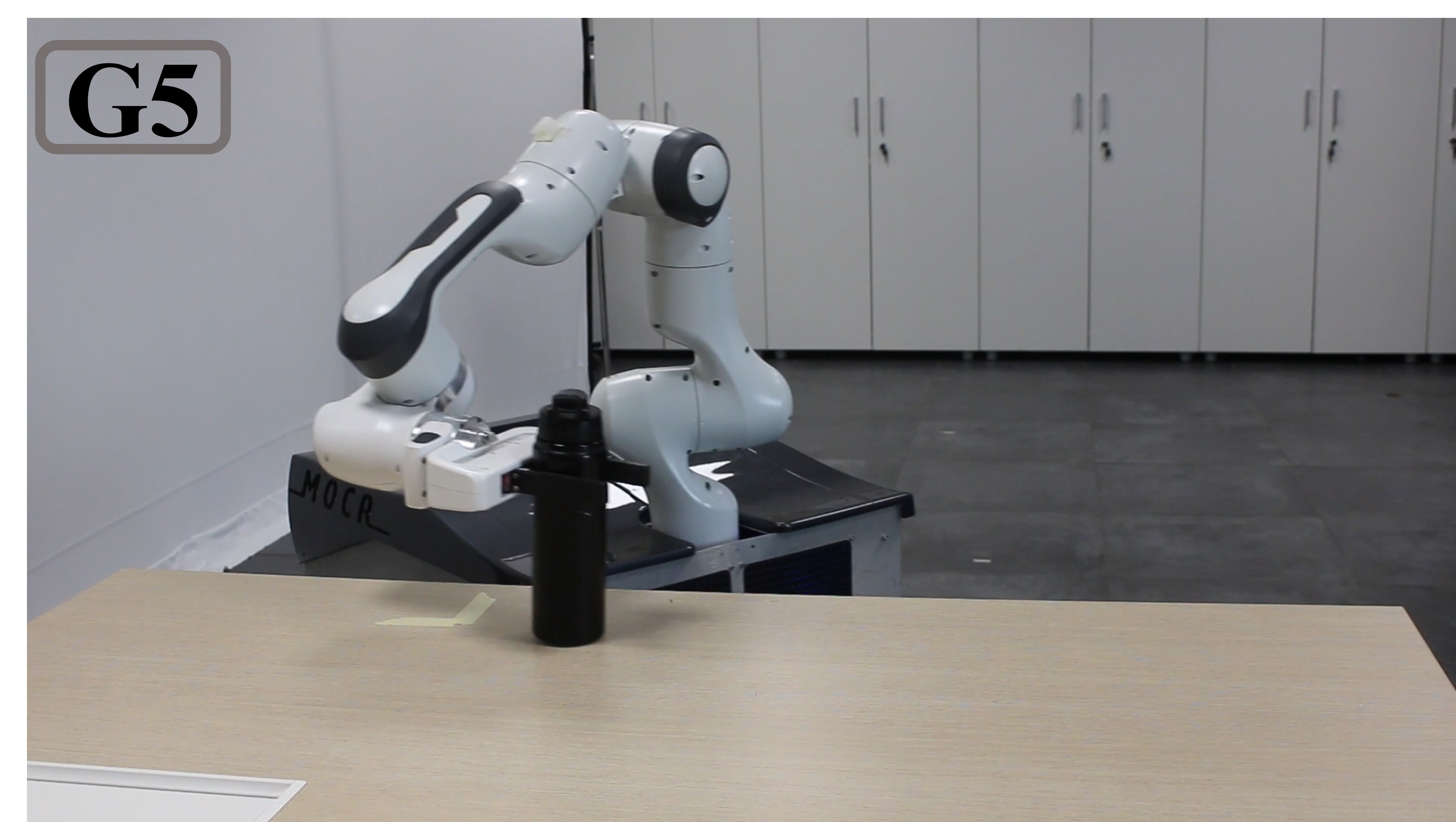}
    \includegraphics[trim=0.0cm 0.0cm 3.0cm 0.0cm,clip,width=0.27\columnwidth]{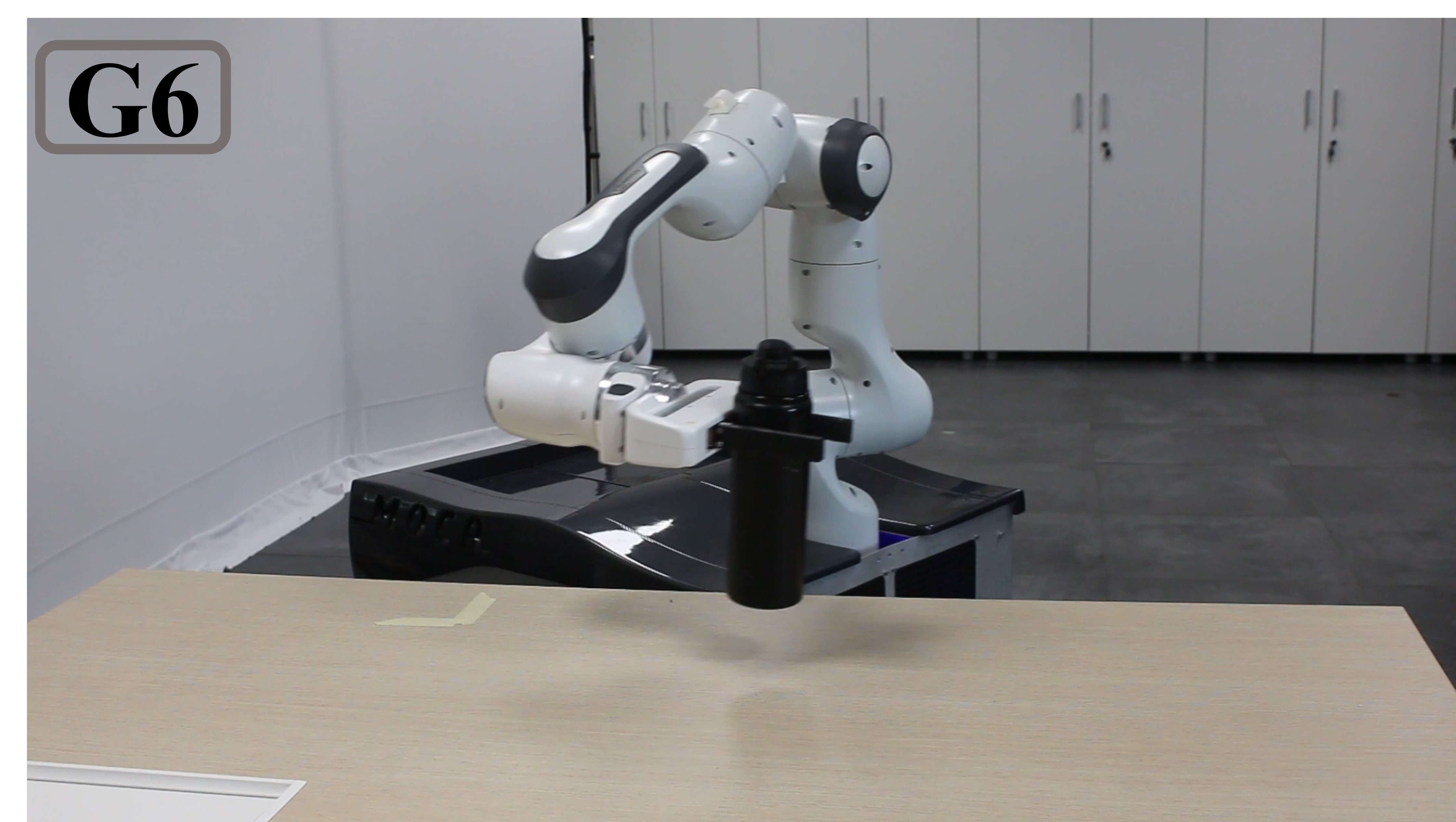}
    \includegraphics[trim=0.0cm 0.0cm 3.0cm 0.0cm,clip,width=0.27\columnwidth]{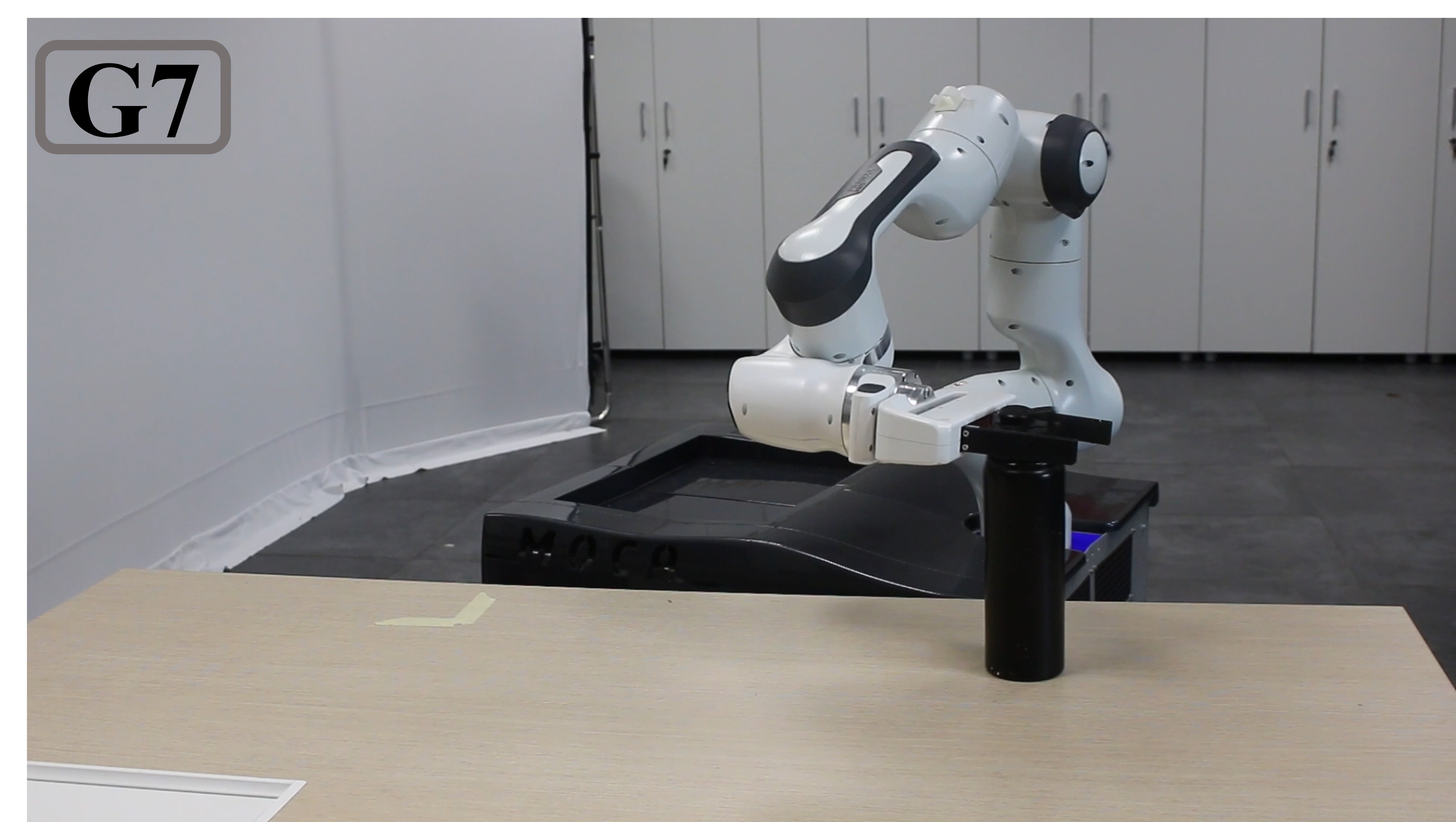}
    \caption{Snapshots of the experiments. (top) Human whole-body demonstration collection for a bottle pick-and-place task. (middle) Experiment 1: autonomous task execution in nominal conditions. (bottom) Experiment 2: autonomous task execution with new randomized relative distance, where the initial EE and base positions were changed.}
    \label{fig:snapshots}
\end{figure*}

The subject was required to pick and place the same bottle $10$ times, according to the following protocol\footnote{The research protocol employed in this study received formal approval from the ethics committee of Azienda Sanitaria Locale (ASL) Genovese N.3, as documented under Protocol IIT\_HRII\_ERGOLEAN 156/2020.}: before each trial, the subject was asked to use his right hand to finish the task and start from the same position (both feet and hand)\footnote{
Based on the definition of $\bm{\Sigma_{EE}}$ and $\bm{\Sigma_{wr}}$, the subject was asked to place his right arm forward at the start (see Fig. \ref{fig:exp_setup} and Fig. \ref{fig:snapshots}). This is done for convenience but not mandatory, and the demonstrated trajectory can be directly applied to other MMs as discussed in Sec.\ref{sec:dis}.
Although we asked the subject to start from the same position, this was not a strict requirement and did not need to be precise.
}. The bottle was located at the same position ($x=3.687~m,y=0.423~m$) on a table (height: $0.723~m$).
Besides, the subject was asked to pick up the bottle naturally while in motion (walking) and place the bottle in a target position ($x=3.73~m,y=0.97~m$) on the table. The overall setting is in the top row of Fig. \ref{fig:snapshots}.  
Although all the movements of the human skeleton were recorded, we only considered the wrist ($\bm{x}_{wr}$, $\dot{\bm{x}}_{wr}$) and pelvis ($\bm{x}_{pe}$) trajectories, as mentioned in Sec.\ref{sec:tra_learn}. Finally, 5 demonstrations were selected to learn the locomotion-integrated pick-and-place task, and the results are presented in the following experiment part.

\begin{figure*}[t]
	\centering
	\includegraphics[trim=0.7cm 4.9cm 0.7cm 4.7cm,clip,width=\linewidth]{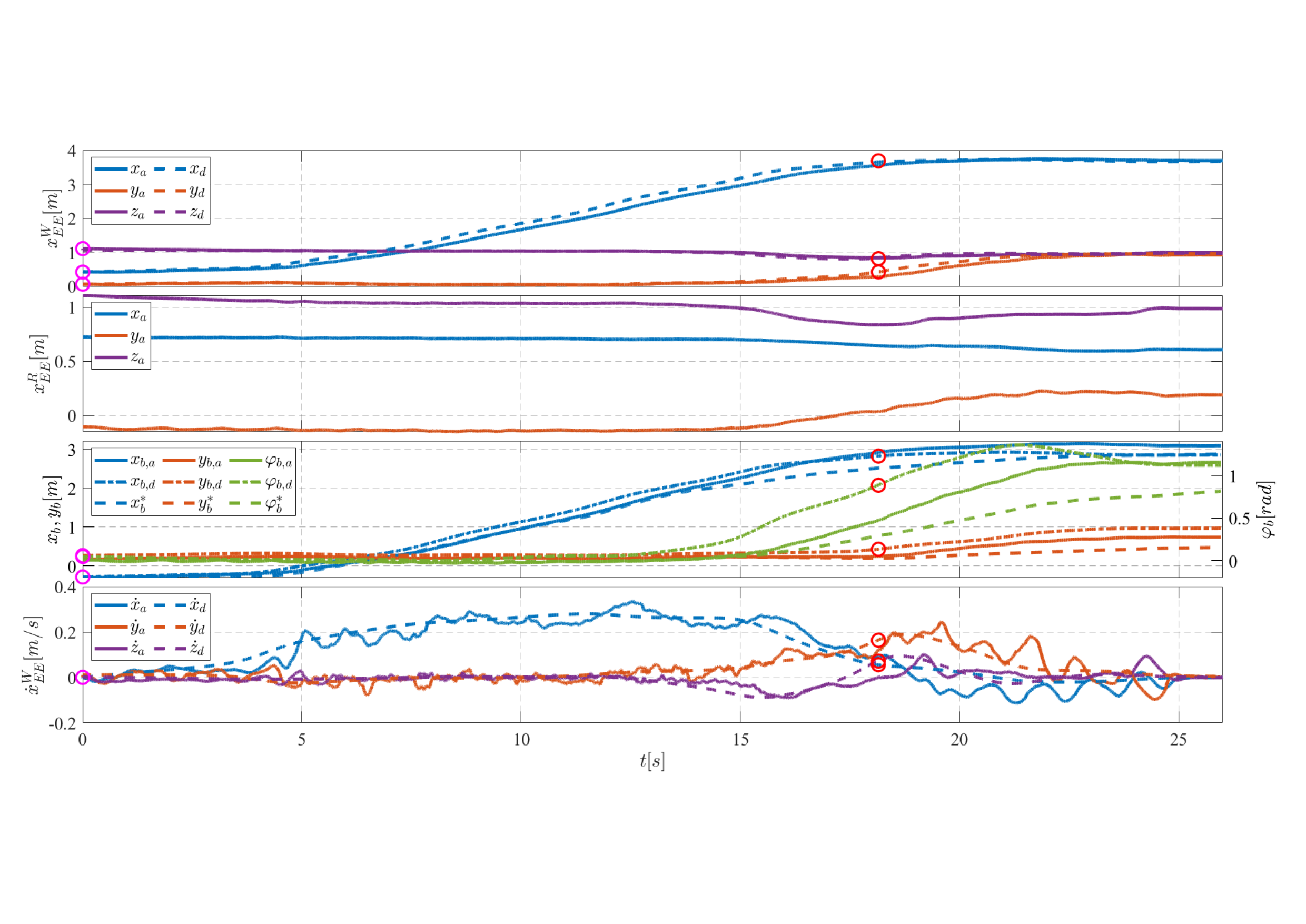}
\caption{Results of the replica task in nominal conditions with MOCA. From top to bottom, first row: EE desired position $\bm{x}_d \in\mathbb{R}^{3}$ and actual position $\bm{x}_a \in\mathbb{R}^{3}$ in the world frame $\bm{\Sigma_W}$; Second row: EE actual position in the robotic arm base frame $\bm{\Sigma_R}$; Third row: desired mobile base learned pose $\bm{q}_{b,d}$, optimal pose $\bm{q}_b^*$ from HQP, and actual pose $\bm{q}_{b,a}$; Fourth row: EE desired velocity $\dot{\bm{x}}_d \in\mathbb{R}^{3}$ and actual velocity $\dot{\bm{x}}_a \in\mathbb{R}^{3}$. At $t=0$, the magenta circles marked the initial states $\bm{x}_d=\ [ 0.417 \quad 0.062\quad 1.107\ ]^T~m$, $\bm{q}_{b,d}=\ [ -0.288~m\quad  0.263~m\quad 0.046~rad\ ]^T$, $\dot{\bm{x}}_d=\ [0\quad 0\quad 0\ ]^T~m/s$; At $t=18.15~s$, the desired grasp point (bottle position) states are highlighted in red circles: $\bm{x}_d=\ [ 3.687 \quad 0.423\quad 0.812\ ]^T~m$, $\bm{q}_{b,d}=\ [ 2.823~m\quad 0.427~m\quad 0.885~rad\ ]^T$, $\dot{\bm{x}}_d=\ [0.056\quad 0.164\quad 0.072\ ]^T ~m/s$.}\label{fig:gmr}
\end{figure*}

\subsection{Experimental Setup}\label{subsec:exp-set}
Here, MOCA is used for the locomotion-integrated pick and placement of the bottle based on learned human skills. The elements $\bm{q}_{b,d}=[x_{pe}\quad y_{pe}\quad \varphi_{pe}]^T=[x_{b,d}\quad y_{b,d}\quad \varphi_{b,d}]^T$ of the pelvis' pose  ($\bm{x}_{pe}$) were learned since the mobile base of MOCA has $3$ DoFs, while only the position and translational speed of wrist's pose and velocity ($(\bm{x}_{wr}\quad \dot{\bm{x}}_{wr})=(\bm{x}_{d}\quad \dot{\bm{x}}_{d})$) were used in all the experiments. Besides, the pelvis frame ($\bm{\Sigma_{pe}}$) should be mapped to the robotic arm base frame ($\bm{\Sigma_{R}}$). Since the commands of the mobile base are described in $\bm{\Sigma_{B}}$, the relative transformation between $\bm{\Sigma_{R}}$ and $\bm{\Sigma_{B}}$ should be compensated (see Fig.\ref{fig:exp_setup}). This offset is $0.3875~m$ for the MOCA in the $x$ direction. The Franka gripper was set to close and open at the bottle position and final point, respectively, which we consider known a priori. The HQP formulation is solved in C++ using ALGLIB QP-BLEIC solver on Ubuntu 20.04.
All experiments were run on a computer with an Intel Core i7-4790S 3.2 GHz $\times$ 8-cores CPU and 16 GB RAM.   

The regularization factor of the KMP were $\lambda=10$. For the HQP  formulation, the gains in CLIK \eqref{eq:min_kyn_clik} were: $\boldsymbol{K_p}=diag\{4,4,4,4,4,4\}$ and $\boldsymbol{K_v}=diag\{1,1,1,1,1,1\}$. We chose higher gains for EE pose and velocity tracking to achieve better accuracy since this is essential to the grasping performance. The control period was $\Delta t=0.001s$. Besides, the selection matrices were $\bm{H}_b=diag\{1,1,1,0,0,0,0,0,0,0\}$ and $\bm{H}_a=diag\{0,0,0,1,1,1,1,1,1,1\}$. 
In this way, the learned mobile base pose tracking (i.e., the first term in \eqref{eq:base_track}) only worked on the optimal base velocity $\bm{q}_{b}^{*}$, and the second term just affected the optimal robotic arm velocity $\bm{q}_{a}^{*}$.
The joint impedance control gains are:
$\bm{K_{q_{a,p}}}=diag\{20,20,20,20,20,20,20\}$ and $\bm{K_{q_{a,v}}}=diag\{8,8,8,8,8,8,8\}$, to have a proper trade-off between tracking accuracy and compliance. 

\subsection{Results and Discussions}
\subsubsection{Human Demonstrations Analysis}\label{subsec:human_demo_analysis}
The snapshots of human demonstrations are presented in the top row of Fig. \ref{fig:snapshots} ($H1-H7$), and the learned trajectories can be found in Fig. \ref{fig:gmr} (i.e., $\bm{x}_d$ in the first row, $\bm{x}_{b,d}$ in the third row, and $\bm{\dot{x}}_d$ in the fourth row, from top to bottom). According to the results, the overall period of the pick-and-place task is $26~s$. Instead of separate and manually designed locomotion, loco-manipulation, and manipulation modes, based on the distance from the CoM to the target object~\cite{ferrari2017humanoid}, there were no pure locomotion and manipulation modes in the learned human behavior. From $0-15~s$, corresponding to $H1-H3$, the human's hand moved slightly in the $z$ direction (from $1.107~m$ to $0.956~m$), and the feet (pelvis) contributed much in the $x$ direction (from $-0.288~m$ to $2.411~m$) with a gradually increased velocity of the EE (from $0~m/s$ to $0.254~m/s$). Starting from $15~s$ ($H3$), the subject started to rotate the body, while the hand movement took the dominant role in all three directions, approaching the target bottle in ($H4$), given the proximity. Meanwhile, the velocity of the EE along the $x$ direction started to decrease, while the value in $y$ and $z$ directions increased for the grasp.
At $t=18.15~s$ , the subject grasped the bottle by hand with non-zero EE velocity $\dot{\bm{x}}_{d}=[\dot{x}_{d}\quad \dot{y}_{d}\quad \dot{z}_d]^T=[0.056\quad 0.164\quad 0.072]~m/s^T$ (red circles in the fourth row of Fig. \ref{fig:gmr}), with a norm of $0.188~m/s$. The dominant element of $\dot{\bm{x}}_{d}$ was along the $y$ direction since the distance between points A and B in Fig. \ref{fig:exp_setup} lies along $y$.
While the subject moved towards the placing position ($H6$), the hand raised gradually, and the feet moved simultaneously. Finally, the human stopped and placed the bottle in the final position ($H7$). 

The overall advantage of the proposed scheme lies in a unique loco-manipulation mode, which encompasses the complex nuances of human behaviors even when performing apparently basic tasks.
Indeed, the largest effect on the EE's movement was changing between the pelvis (base) and the arm in different phases. First, the pelvis moved to get close to the bottle. Then, the arm took the main role and picked the bottle. After that, the pelvis and the arm's motion were synergical until the final point.
This observed behavior shared some similarities with~\cite{ferrari2017humanoid}, where the locomotion, loco-manipulation, and manipulation modes were chosen based on the distance from the target object.
It is worth addressing that we focus on learning the coordination movement between the mobile base and the robotic arm in locomotion-integrated tasks rather than grasping.

\subsubsection{Learned Skills Replica with MOCA}\label{sec:repeat}

We used MOCA to replicate the learned pick-and-place autonomously. Firstly, the task was successfully achieved using the same initial settings (i.e., initial pose and bottle location), whose results are shown in the middle row ($R1-R7$) of Fig. \ref{fig:snapshots} and Fig. \ref{fig:gmr}, respectively. 
From the initial configuration ($R1$ in Fig. \ref{fig:snapshots}), MOCA moved towards the bottle ($R2$) and started to rotate the base (as the human did), approaching the bottle ($R3$).
Then MOCA grasped the bottle ($R4,R5$) and placed it in the desired position ($R6,R7$). 
Differently from the human's grasping, it is worth noting that the EE dragged the bottle on the table for a short period (between $R4$ and $R5$) while closing the gripper. This is because the closing velocity is significantly slower than human fingers, even under maximum conditions. Thanks to the learned EE velocity, the bottle did not fall,
enabling a non-zero velocity grasp.
\begin{figure*}[t]
	\centering
	\includegraphics[trim=0.7cm 4.9cm 0.7cm 4.7cm,clip,width=\linewidth]{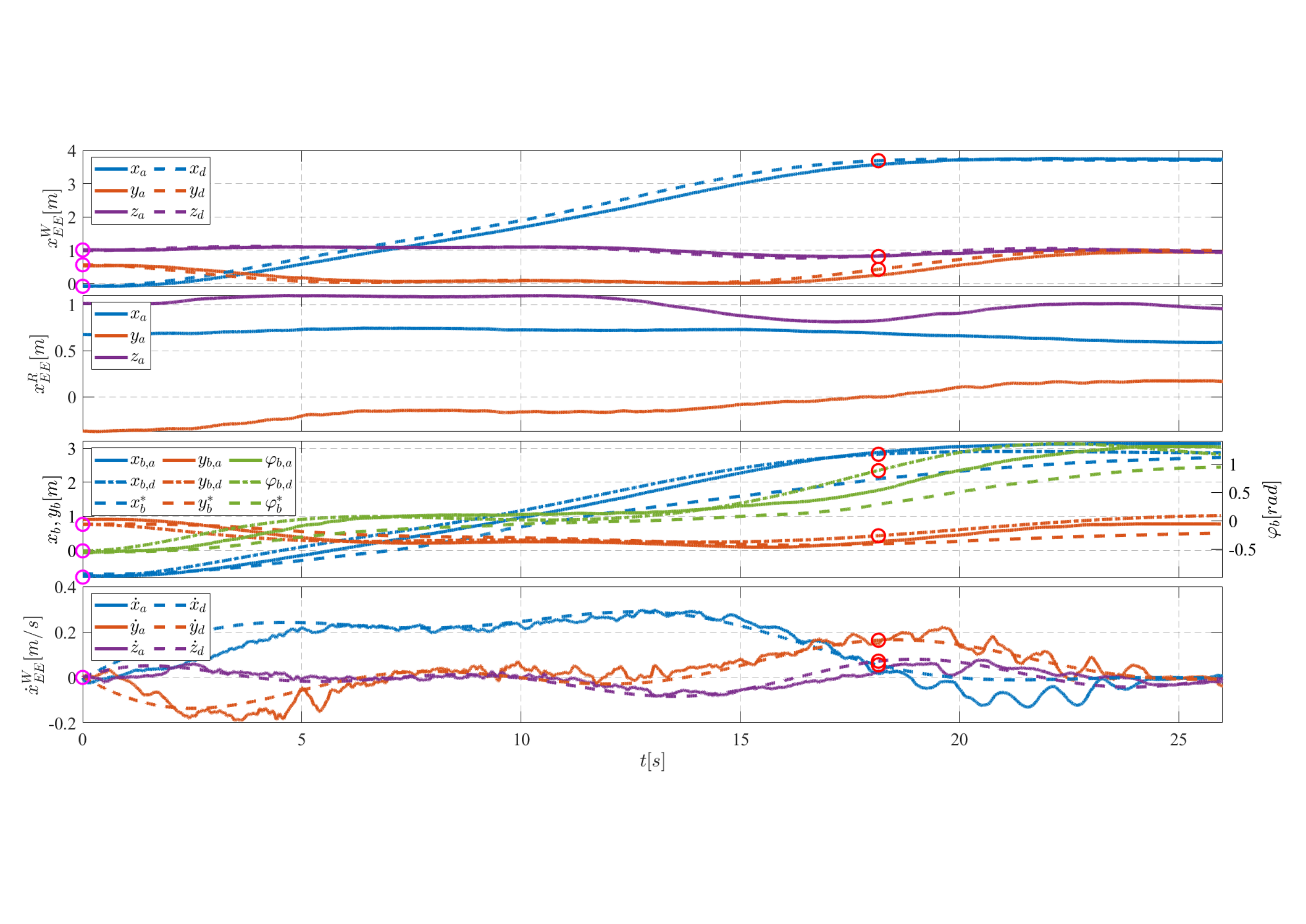}
	\caption{Results of autonomous task execution experiments in generalized conditions with MOCA. At $t=0$, the magenta circles mark the new initial states $\bm{x}_d=\ [ -0.0827 \quad 0.5623\quad 1.007\ ]^T~m$, $\bm{q}_{b,d}=\ [ -0.7883~m\quad  0.7634~m\quad -0.53~rad\ ]^T$; At $t=18.15~s$, the same desired grasp point (bottle position) are highlighted in red circles. The EE desired initial and grasp velocities are the same as in Fig. \ref{fig:gmr}.}
	\label{fig:kmp}
\end{figure*}

The tracking performances with respect to the learned EE position and velocity ($\bm{x}_{EE}^W,\dot{\bm{x}}_{EE}^W$) in world frame ($\bm{\Sigma_{W}}$) are shown in the first and fourth row of Fig. \ref{fig:gmr}, respectively. As the highest priority level in the SoT of the HQP, the overall performance was satisfactory, with the Root
Mean Squared Error (RMSE)\footnote{$RMSE_{\bm{x}_d}=\sqrt{\frac{1}{N} \sum_{i=1}^N (\bm{x}_{d,i}-\bm{x}_{a,i})^2}$, with total number $N$ of samples.} $RMSE_{\bm{x}_d}=\ [ 0.1428 \quad 0.0505\quad 0.0250\ ]^T~m$ and $RMSE_{\dot{\bm{x}}_d}=\ [ 0.0415 \quad 0.0353\quad 0.0241\ ]^T~m/s$.  
As for the learned mobile base's pose tracking (third row of Fig. \ref{fig:gmr}), the performance was inferior with respect to the EE pose tracking as expected, with the RMSE between the learned base pose $\bm{q}_{b,d}$ and the optimal pose $\bm{q}_{b}^*$ generated by the HQP $RMSE_{\bm{q}_{b,d}}=\ [ 0.1912~m \quad 0.2297~m\quad 0.3132~rad\ ]^T$, which is higher than $RMSE_{\bm{x}_d}$, as expected. Since the mobile base pose tracking was set as the second layer in SoT, the tracking accuracy was sacrificed to ensure the first layer task performance (i.e., the EE pose and velocity tracking).
The RMSE between the optimal base pose $\bm{q}_{b}^*$ and the actual pose is $\bm{q}_{b,a}$ $RMSE_{\bm{q}_{b}^*}=\ [ 0.1875~m \quad 0.1149~m\quad 0.1735~rad\ ]^T$ which indicates the trajectory tracking accuracy of the lower level velocity controller of the base.
Based on the proposed approach, the robot learned the human demonstrated skills successfully, achieving pick-and-place tasks.

\subsubsection{Learned Skills Generalization with MOCA}

Commonly, the robot starts from different initial positions in pick-and-place tasks. Owing to the generalization capacity of KMP, we can generalize the learned skills to new settings. Specifically, the initial EE position was changed from $(0.417,\quad 0.062,\quad 1.107)~m$  to $(-0.0827,\quad 0.5623,\quad 1.007)~m$. This affects the relative path generated; hence, the bottle position remains the same. The desired grasp time is again $t=18.15~s$. Being the EE velocity and base pose in the output variable vector $\bm{\xi}$, we should choose proper values for the new desired points. Obviously, the EE velocity at the initial point is zero, and the demonstrated contact velocity should be the same at the new bottle position to ensure a successful grasp. For the base pose, the values were changed from $(-0.288~m,\quad 0.263~m\quad 0.046~rad)$ to $(-0.7883~m,\quad 0.7634~m\quad -0.53~rad)$ for the initial point.
Therefore, the new desired initial point was inserted into the reference trajectory, and a new trajectory was generated by \eqref{eq:klmean}. The generalized trajectory was sent to MOCA, achieving a successful pick-and-place task. 
The task processes and results are shown in the bottom row in Fig. \ref{fig:snapshots} ($G1-G7$) and Fig. \ref{fig:kmp}, respectively. 
Although the initial EE and base position changed significantly, MOCA showed similar behavior to the \textit{replica case} in Sec. \ref{sec:repeat}. From the new initial point ($G1$, clearly different from $R1$), MOCA moved close to the bottle ($G2$). Then, the mobile base was rotated, and the arm approached the bottle's new position ($G3$). The EE and the bottle came into contact, and the bottle was grasped in ($G4, G5$). Eventually, MOCA transported the bottle to the desired point ($G6$) and placed it ($G7$) on the table. Due to the gripper not being closed immediately, the bottle was briefly dragged on the table's surface as before. However, MOCA still achieved stable grasping, thanks to the hierarchical velocity profile generated online.

According to Fig. \ref{fig:kmp}, the generalized reference trajectory had a similar shape concerning the \textit{replica case}. 
Magenta circles highlight the original and new desired initial points, and the desired grasping point is marked with red circles in Fig. \ref{fig:gmr} and Fig. \ref{fig:kmp} separately. 
The overall tracking performances of both EE pose/velocity are $RMSE_{\bm{x}_d}^{'}=\ [  0.1600 \quad 0.0867\quad 0.0366\ ]^T~m$ and $RMSE_{\dot{\bm{x}}_d}^{'}=\ [ 0.0496 \quad 0.0410\quad 0.0264\ ]^T~m/s$, while for the mobile base $RMSE_{\bm{q}_{b,d}}^{'}=\ [ 0.5102~m \quad 0.2682~m\quad 0.3652~rad\ ]^T$ and $RMSE_{\bm{q}_{b}^*}^{'}=\ [ 0.4732~m \quad 0.1882~m\quad 0.2587~rad\ ]^T$. It is worth mentioning that the RMSE of the base tracking increased, given the more complex geometry of the trajectory (third row of Fig. \ref{fig:gmr} and Fig. \ref{fig:kmp}), but the EE's RMSE remained similar to the \textit{replica case}, showing how the secondary task is sacrificed at the expenses of the primary one, resulting in a successful task.

\subsubsection{Discussions}\label{sec:dis} according to the presented results, the human demonstrated loco-manipulation skills of a pick-and-place task were successfully transferred to MOCA
based on the proposed combined learning and optimization approach. Moreover, the learned whole-body trajectory can be generalized to unforeseen settings by KMP. The proposed learning method generates a single smooth whole-body trajectory for the overall task and does not need to define different points where to switch between motion modes, compared with the most relevant research~\cite{ferrari2017humanoid}.
For the controller side, the proposed HQP formulation can execute the overall whole-body trajectory smoothly without stopping and changing parameters, resulting in the main contributor to the EE motion being switched between the robotic arm, mobile base, or both. This constitutes an advancement with respect to the weighted whole-body cartesian impedance controller in our previous paper~\cite{wu2021unified}, which needed a transition phase ($10~s$) to change the weighting parameters to switch between three motion modes. Therefore, the proposed HQP controller improves execution time and efficiency.

The pick-and-place task showed the capacity of the proposed learning and optimization framework, which can be directly applied to other loco-manipulation scenarios, for example, the door-opening task in~\cite{lfdmm2017tim}. Although we evaluated the demonstrated pick-and-place skills on MOCA, unlike the IRM-based methods for a specific mobile manipulator, it is possible to apply the learned skills to other MMs, such as TIAGo++ in~\cite{jauhri2022robot}, PR2 in~\cite{lfdmm2017tim}, NAO humanoid in~\cite{ferrari2017humanoid}, and even legged MMs~\cite{zimmermann2021go}, as long as the learned EE trajectory 
is achievable for the MM. 
Furthermore, owing to the hierarchical design in the proposed HQP, the learned EE trajectory was tracked accurately, and the base pose was followed as a secondary priority, which allowed the transfer of the human whole-body mobile manipulation skills to a robot with different geometry.    

\section{Conclusions}\label{sec:conclusion}
The combined learning and optimization approach developed in this paper successfully transferred human-demonstrated pick-and-place skills to MOCA, a MM with a different geometry from that of the human subject. Thanks to KMPs, the learned skill was generalized to a new initial position with the same contact velocity. In this way, MOCA grasped the bottle with the learned human velocity profile. The learning part of the proposed approach generated the whole-body motion for loco-manipulation tasks, considering the non-zero contact velocity of the EE.
Receiving the whole-body reference trajectory from the KMP, the HQP part of the proposed approach generated the feasible and optimal joint-level commands at each time step. Owing to the hierarchical design, the proposed approach allowed the transfer of the human demonstrated skills to the MM.

Future works will include learning the spatial relationship between the human wrist and pelvis in loco-manipulation tasks. The new desired EE pose can be easily found according to the task requirements, but defining the corresponding mobile base pose is not intuitive. In this work, the bottle was static, and the position was pre-defined. Considering the real-time motion of the bottle to the proposed approach, a closed-loop reactive learning approach is another appealing future development.

\bibliographystyle{IEEEtran}
\bibliography{biblio}

\end{document}